\documentclass[preprint,5p,times,twocolumn]{elsarticle}

\usepackage{amssymb}
\usepackage{amsfonts}
\usepackage{amsmath}
\usepackage{algorithm}
\usepackage{algorithmic}
\usepackage{multirow}
\usepackage[colorlinks,
            linkcolor=blue,
            anchorcolor=blue,
            citecolor=blue]{hyperref}

\journal{Expert Systems With Applications}

\begin{document}

\begin{frontmatter}



\title{Learning Semantic-Aware Threshold for Multi-Label Image Recognition with Partial Labels}


\author[1]{Haoxian Ruan\fnref{equal}}
\ead{HowxRuan@mail2.gdut.edu.cn}

\author[1]{Zhihua Xu\fnref{equal}}
\ead{zihua@mail2.gdut.edu.cn}

\author[1]{Zhijing Yang}
\ead{yzhj@gdut.edu.cn}

\author[2]{Guang Ma}
\ead{eddiemag@163.com}

\author[1]{Jieming Xie}
\ead{JmXie39@gdut.edu.cn}

\author[3]{Changxiang Fan}
\ead{fanchangxiang@gdaas.cn}

\author[1]{Tianshui Chen\corref{cor1}}
\ead{tianshuichen@gmail.com}

\fntext[equal]{Haoxian Ruan and Zhihua Xu contribute equally to this work and share first authorship.}

\address[1]{School of Information Engineering, Guangdong University of Technology, Guangzhou, China}
\address[2]{Department of Computer Science, University of York, UK}
\address[3]{Institute of Facility Agriculture, Guangdong Academy of Agricultural Sciences (IFA, GDAAS)}

\cortext[cor1]{Corresponding author}


\begin{abstract}

Multi-label image recognition with partial labels (MLR-PL) is designed to train models using a mix of known and unknown labels. Traditional methods rely on semantic or feature correlations to create pseudo-labels for unidentified labels using pre-set thresholds. This approach often overlooks the varying score distributions across categories, resulting in inaccurate and incomplete pseudo-labels, thereby affecting performance. In our study, we introduce the Semantic-Aware Threshold Learning (SATL) algorithm. This innovative approach calculates the score distribution for both positive and negative samples within each category and determines category-specific thresholds based on these distributions. These distributions and thresholds are dynamically updated throughout the learning process. Additionally, we implement a differential ranking loss to establish a significant gap between the score distributions of positive and negative samples, enhancing the discrimination of the thresholds. Comprehensive experiments and analysis on large-scale multi-label datasets, such as Microsoft COCO and VG-200, demonstrate that our method significantly improves performance in scenarios with limited labels.

  
\end{abstract}




\end{frontmatter}


\section{Introduction}
Multi-label image recognition (MLR), which assigns multiple labels for each instance simultaneously, is a practical task in computer vision that has received much attention \cite{tian2023causal, tang2022image, wang2023image, sun2014multi}. It is widely used in many application scenarios ranging from content-based image retrieval \cite{cheng2005semantic, li2010technique, zhang2021instance, lai2016instance}, recommendation systems \cite{carrillo2013multi, darban2022ghrs, zheng2014context} to human attribute analysis \cite{chen2021cross, pu2021expression, wu2019instance, lima2014multi}. Thanks to the development of deep neural networks, significant progress has been made in this task in recent years \cite{ridnik2021asymmetric, chen2021learning, wu2020adahgnn}. Despite the excellent performance of these methods, one should see that they heavily rely on large-scale and high-quality datasets for learning discriminative features. With the number of samples and categories increase, collecting clean and complete labels for each image is very laborious and time-consuming, which makes annotating a large-scale multi-label benchmark a difficult task. In practical application scenarios, it is customary to address the challenge posed by expanding datasets, wherein both the volume of individual samples and the diversity of categories continuously increase over time. Thus, a more flexible strategy for data annotation and corresponding model learning scheme are crucial. Collecting a multi-label dataset with partial annotation is an alternative and feasible strategy to solve the above problems. For a partially annotated image, only a subset of positive and negative labels are known, while the others are unknown. It’s obvious that the strategy of partial annotating is much easier and expandable and is more suitable for actual scenarios. In this paper, we aim to address the task of learning multi-label recognition models with partial labels, allowing them to maintain excellent performance with limited supervision information.

Current methods mainly model multi-label as a multiple-binary classification task. In MLR-PL task, a straightforward approach is either to ignore unknown labels or to simply classify them as negative labels \cite{kundu2020exploiting, huynh2020interactive, sun2017revisiting, joulin2016learning}. However, ignoring the unknown labels may lead to data loss and reduce generalization capability of the model. Meanwhile, treating unknown labels as negative may incurs some noisy labels and leads to potential misdirection during training. 

To address these issues, some methods based on pseudo-labels have been proposed previously \cite{chen2022heterogeneous, chen2022structured}, where known labels are utilized for model training first, then the trained model can generate pseudo-labels for those unknown labels whose prediction scores are above pre-defined thresholds, and finally these pseudo-labels are seen as known labels for training. 
In these methods, the selection of suitable thresholds is crucial for the quality of generated pseudo-labels. The mainstream methods mainly consider a uniform threshold for all categories \cite{rizve2021defense, sohn2020fixmatch}, or a customized threshold adjusting scheme \cite{li2022rethinking, zhang2021flexmatch, chen2022heterogeneous}. The former category of works \cite{rizve2021defense, sohn2020fixmatch} typically utilize a rather high threshold throughout the training process. Although it can ensure the quality of pseudo-labels, the high threshold may ignore the inter-class gap among different categories and lead to low data utilization in the early training stages. Also, searching for a suitable threshold for each dataset is very time-consuming and resource-intensive. The latter branch of methods \cite{li2022rethinking, chen2022heterogeneous, chen2022structured, zhang2021flexmatch} propose gradually adjusting thresholds during training. One potential limitation of these methods is that they utilize a uniform threshold for all categories \cite{chen2022heterogeneous, chen2022structured}, where the differences between categories are still ignored. Other works \cite{li2022rethinking, zhang2021flexmatch} propose class-specific thresholds. However, these methods either fail to consider learning statuses of different categories \cite{li2022rethinking}, or lack direct connection to the model output \cite{zhang2021flexmatch}. 

To determine the optimal thresholds for generating high-quality pseudo-labels, we need to consider the unique characteristics of different categories and the dynamic aspects of the learning process simultaneously. As shown in Figure \ref{Fig:Exp1}, the output distributions of different categories vary, which illustrate the intrinsic differences in different categories and the information about model's learning status. Therefore, it is necessary to set class-specific, self-adaptive thresholds during the training process. To estimate thresholds that meet these requirements, we investigate a reasonable assumption: the model exhibits consistency in feature extraction and prediction for samples within the same category, even when the labels are inconsistent. This consistency is key in transferring information from known to unknown labels within a category. For a more detailed analysis, we delve into the examination of correlation between the model output distributions of known and unknown samples. As shown in Figure \ref{Fig:Exp2}, there exists strong similarity between these distributions despite disparities in sample sizes. This observation provides support for the approximation of unknown distributions by using the known distributions within the same category. Consequently, we can obtain statistical information about the distribution of unknown samples, and use this information to determine reasonable thresholds for different categories.

Based on the above analysis, we introduce a novel Semantic-Aware Threshold Learning (SATL) method, which estimates the output distributions of unknown samples by utilizing information from known samples, and subsequently determines a threshold for each category. 
Formally, we introduce the Semantic-Aware Threshold Estimation (SATE) module, which leverages existing labeled data to estimate the distributions of confidence scores for both positive and negative instances within each category. Given that features belonging to the same category share semantic similarity, it is reasonable to derive the distribution estimation from known labels to unknown ones. Consequently, we can establish an ideal category-specific threshold, informed by the understood distributions of positive and negative instances. 
Meanwhile, for difficult categories, the output distributions of positive and negative samples may overlap significantly,leading to conservative pseudo-label predictions and a reduction in the number of recalled labels. To address this challenge, we further introduce a differential ranking loss (DRL) to widen the gaps between positive and negative samples, which aims to enhance the discriminative ability of the estimated thresholds.

The contributions of our work can be summarized as follows. First, we introduce the Semantic-Aware Threshold Learning (SATL) algorithm, which innovatively computes score distributions for both positive and negative samples across categories, enabling the determination of precise category-specific thresholds. \textcolor{black}{In comparison with previous methods that apply a global or heuristic threshold, the SATL algorithm jointly considers inter-class differences and model training dynamics, which enhances pseudo-label mining.} Second, we design a differential ranking loss (DRL) to significantly widen the gap between the score distributions of positive and negative samples, thus improving the discriminative power of the model. Finally, we validate the effectiveness of the proposed algorithm by extensive experiments on major multi-label datasets such as Microsoft COCO and Visual Genome 200.

\section{Related Works}

\subsection{MLR with Complete Labels}

Multi-label image recognition (MLR) has received much attention in recent years. It is more practical and necessary than its single-label counterpart since images of daily life generally contain multiple objects varying from different categories. To solve this task, many works have been dedicated to obtain discriminative local regions with objects by using object proposal methods \cite{wei2015hcp, yang2016exploit} or attention mechanisms \cite{gao2021learning, chen2018recurrent, wang2017multi}. 
Another line of works considered utilizing category dependencies to regularize the training of models and provide more information.
Some works introduced time series models such as recurrent neural network (RNN) or long-short term memory network (LSTM) to capture category dependencies implicitly \cite{wang2016cnn, wang2017multi, chen2018recurrent}.
Other works proposed to explicitly model the correlations between the categories by introducing a graph neural network (GNN) \cite{ou2020multi, zhang2023graph, chen2024dynamic, chen2024learning}.
Chen et al. \cite{chen2019learning} proposed a semantic-specific graph representation learning (SSGRL) framework using a graph propagation network to obtain decoupled semantic-aware features, which achieved state-of-the-art performance on several multi-label datasets. In \cite{chen2019multi}, an inter-dependent object classifier based on a graph convolutional neural network (GCN) was proposed to explicitly model the correlations between different categories. 

Despite achieving remarkable performance, these methods rely on deep neural networks (DNN) that require large-scale and clean datasets (e.g., MS-COCO, Visual Genome and Pascal VOC) to learn discriminative data representations. However, it is very time-consuming and labor-intensive to annotate a complete list of labels for each image, which makes collecting large-scale and complete multi-label datasets more difficult and unscalable. Therefore, we focus on the MLR-PL task, which exhibits broader applicability in practical contexts.

\subsection{MLR with Partial Labels}
Due to the difficulty of annotating complete multi-label datasets, the task of multi-label image recognition with partial labels (MLR-PL) is getting more and more attention \cite{pu2024dual, ruan2024learning, ma2019label, wang2021pico}, in which only part of labels are known in each image. Earlier works \cite{joulin2016learning} addressed this task by simply regarding the unknown labels as negative labels and following the fully supervised training process. Intuitively, the performance of these methods may drop significantly, since they introduce large amount of false negative labels and cause misleading direction for learning. Other works treated MLR as an independent binary classification task for each category \cite{tsoumakas2007multi}. However, these methods fail to consider the correlations between labels which are essential in MLR task. 

To solve this problem, current works tend to introduce pseudo-labels for the complement of unknown labels. For example, Durand et al. \cite{durand2019learning} proposed the partial-BCE loss normalized by the proportion of known labels to perform supervised training. Also, they used pre-trained model to generate pseudo-labels through the curriculum learning strategy. In \cite{kim2022large}, Kim et al. proposed to recognize positive labels by observing fluctuations of loss for each sample due to the memorization effect in DNN. Based on pseudo-labeling framework, some other works propose exploiting label dependencies to transfer knowledge from known labels to unknown labels. In \cite{chen2022structured}, Chen et al introduced a structured semantic transfer framework (SST) with an intra-image transfer module to learn label co-occurrence relationships and a cross-image transfer module to capture semantic similarities, both of which generate pseudo-labels for training. In \cite{chen2022heterogeneous}, the HST framework further proposed a learning algorithm for the dynamic adjustment of threshold.

Following mainstream methods, our SATL framework is based on the pseudo-labeling framework. In this framework, the selection of thresholds, which filter out potential pseudo-labels, is crucial for the quality of the generated pseudo-labels. Different from previous methods, our SATL proposes a novel threshold selection method that leverages the output distribution of known samples, accounting for the differences between categories and model learning status.

\subsection{Threshold Learning}
In pseudo-labeling methods, a suitable confidence threshold is crucial for generating high quality pseudo-labels. Some methods considered using a fine-tuned fixed threshold to filter out reliable pseudo-labels \cite{sohn2020fixmatch, rizve2021defense}. However, searching for a suitable threshold is time-consuming and resource-intensive. Meanwhile, keeping a fixed threshold ignores the different learning status and the intrinsic gapes of different categories. 
Other methods considered to introduce dynamic threshold adjustment. The SST framework \cite{chen2022structured} simply adopt a linear decay scheme for the adjustment of a global threshold until it reaches its minimum. Furthermore, in \cite{chen2022heterogeneous}, chen et al. proposed a DTL method to adjust global threshold throughout the training process. However, These methods fail to capture the learning status of model and the differences between categories. Some other works introduce class-specific thresholds with varying motivations. FlexMatch \cite{zhang2021flexmatch} proposed that different classes should have different local thresholds. While the learning difficulties of different classes is taken into account, these thresholds are estimated by multiplying the variance with a fixed threshold, which is actually misaligned with the meaning of model prediction scores. In \cite{zhang2023refined}, Zhang et al. proposed a category-aware adaptive threshold estimation module to alleviate the problem of imbalanced class distributions, which assign high thresholds to majority categories and low thresholds to minor categories. Similarly, Li et al. \cite{li2022rethinking} proposed to estimate class-specific thresholds with the knowledge of confidence level for each positive sample and the number of positive instances. However, these methods fail to consider the dynamic aspect of model learning status, making them less flexible and resulting in poor performance.

Different from the above methods, we stress the importance of introducing class-specific thresholds in pseudo-label learning methods. Additionally, we present an in-depth analysis of the correlations between the output distributions of known and unknown samples within the same category, and find there exists strong similarity between these distributions. Therefore, we propose estimating the statistical information about the distributions of unknown samples, and leverage this information to determine suitable class-specific thresholds, where both the intrinsic differences between categories and the learning status of model are taken into account.

\begin{figure}[t] 
	\centering 
	\includegraphics[width=1.0\linewidth]{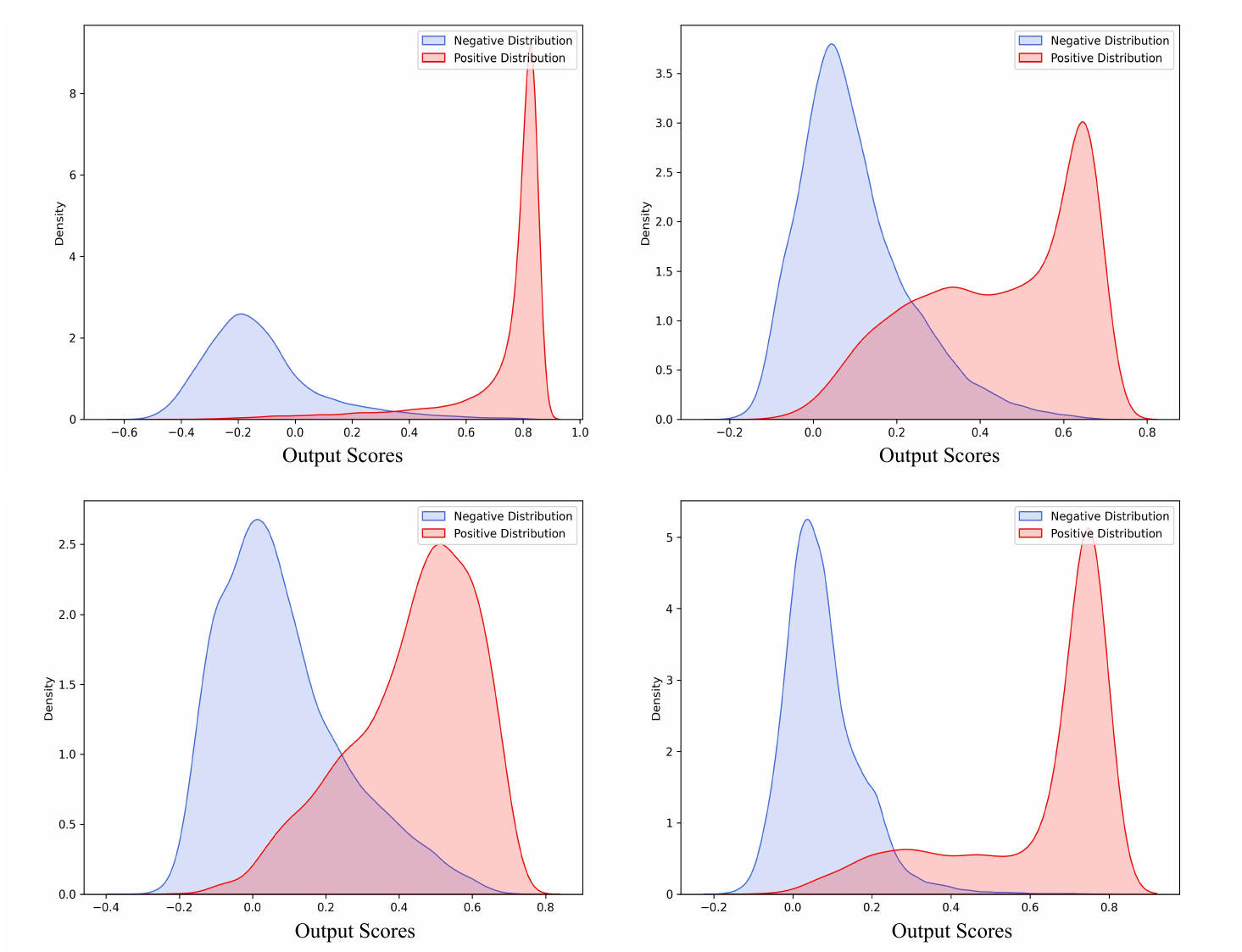}
	\caption{Positive and negative distributions of model's output scores in four different categories, where the red part stands for positive distribution while the blue part is the negative distribution. These distribution curves exhibit unique patterns, emphasizing the necessity of utilizing class-specific thresholds.}
	\label{Fig:Exp1}
\end{figure}

\begin{figure}[t] 
	\centering 
	\includegraphics[width=1.0\linewidth]{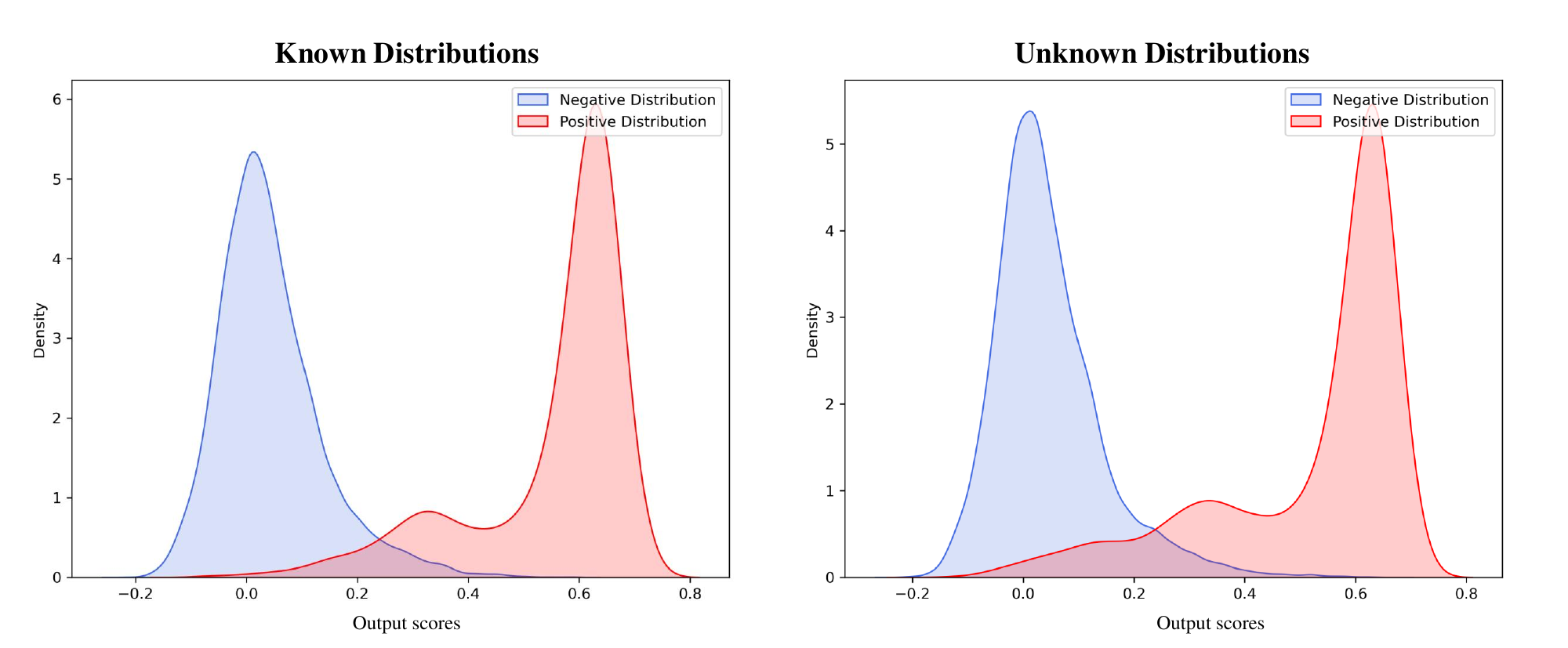}
	\caption{Known and unknown distributions of model's output scores within the same category. The alignment of these distribution curves suggests the possibility to approximate unknown distributions by using the known ones.}
	\label{Fig:Exp2}
\end{figure}

\section{Motivation}
As discussed above, to generate plausible pseudo-labels, we should consider the class-specific threshold due to the different learning status and domain gap among categories. Also, since visual features within the same category exhibit similarity, we make a reasonable assumption that the known and unknown distributions of predictions would align closely despite label inconsistencies. Therefore, the output distribution of unknown samples can be approximated by using known samples. To see these notions more directly, we further conduct experiments and present visualization results for a more direct understanding of these assumptions.

Formally, we conduct a training process with partial labels. To simulate the conditions in partial label setting, we randomly set some labels as unknown for each image, which are invisible to model while proceeding the training process. Then, we follow prior work \cite{chen2022structured} to implement both our feature extractor and pseudo-label generator. For each image $I^n$, we utilize the feature extractor $\phi$ to obtain its feature map, followed by the pseudo-label generator $\varphi$, which maps the feature into a pseudo confidence distribution $s^n$. This procedure can be formulated as follows:
\begin{equation}\label{eq1}
    s^n=[s^n_1, s^n_2, ..., s^n_C]=\varphi \circ \phi(I^n)
\end{equation}
where $s^n_c$ represents the confidence score of the $c$-th category and $C$ is the number of categories. We then classify $s^n_c$ based on its label. If the label is unknown to the model, $s^n_c$ is assigned as an unknown sample; otherwise, it is considered a known sample. Additionally, we further divide $s^n_c$ into positive samples if the label is positive or negative samples if the label is negative. After processing the entire dataset, these samples are aggregated by category and we can visualize their distributions. Consequently, the selection of pseudo-labels in each category can be transformed into a binary classification task using a threshold.

We first consider the distributions of positive and negative samples across various categories, which are shown in Figure \ref{Fig:Exp1}. These distribution histograms exhibit distinct patterns among different categories, which reveal inherent category-specific gaps. Also, the distribution curves provide extra information about the learning difficulty within each category. A significant overlap between the positive and negative sample distributions suggests a challenging category since they are mixed up, whereas a clear separation indicates an easier category. These results indicate the need for class-specific thresholds, considering the differences between categories and their learning difficulties.

We then examine the distributions of known and unknown samples within the same category. Although the unknown labels are invisible to the model, we can still access their ground truth values. Thus, we can visualize the positive and negative distributions of known and unknown samples in a specific category, as shown in Figure \ref{Fig:Exp2}. These two distributions are similar despite variations in the sample sizes. Therefore, the distribution of unknown samples can be reasonably approximated by using the known samples within the same category. Even though the true values of unknown labels are inaccessible to the model, it can still get information from the prediction confidence $s^n_c$. 

In conclusion, setting class-specific thresholds is necessary due to the inherent disparities between categories. For a specific category, a suitable threshold should be determined based on the statistical information on the unknown samples. To achieve this, we can estimate the unknown distribution from the known samples and thereby determine the optimal threshold.

\section{Methods}
\begin{figure*}[!ht] 
	\centering 
	\includegraphics[width=1.0\linewidth]{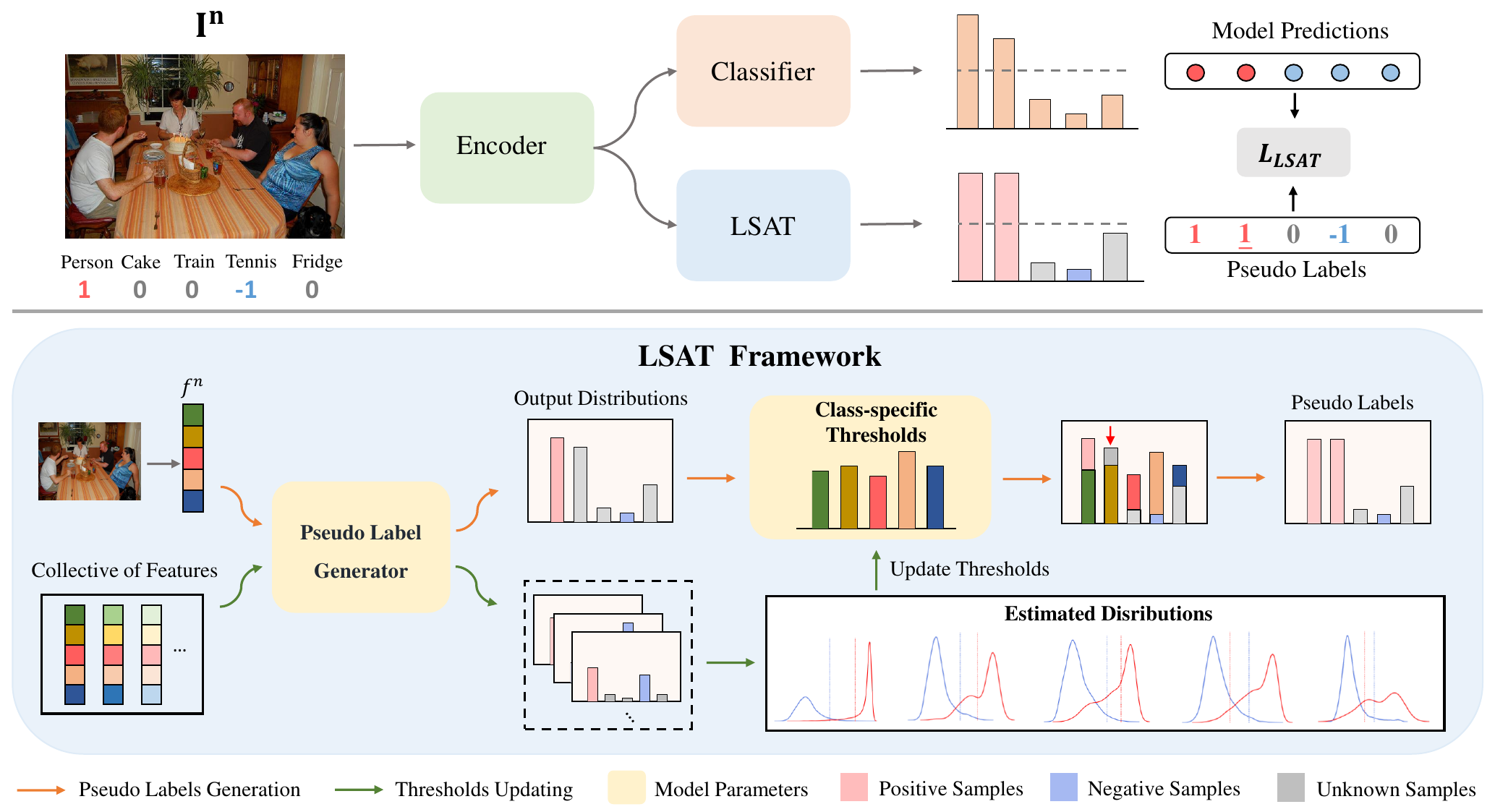}
	\caption{An overall illustration of the proposed learning semantic-aware threshold (SATL) framework. The upper part is the overall pipeline that consists of the pseudo-labeling framework in MLR-PL and the SATL module. Image features and known labels are fed into SATL module to generate pseudo-labels for extra supervision information. The lower part is the detailed SATL module. The output distributions of pseudo-label generator are used for estimation of ideal class-specific thresholds, which are later used for the updating of model's class-specific thresholds. The model then filters out new pseudo-labels using the updated thresholds and outputs the distribution of the input image.}
	\label{Fig:SATL}
\end{figure*}
In this section, we introduce the proposed Semantic-Aware Threshold Learning (SATL) framework that identifies optimal class-specific threshold for each category. The proposed framework consists of two complementary components: the Semantic-Aware Threshold Estimation module (SATE) and the Differential Ranking Loss (DRL), both of which can be easily integrated into general pseudo-labeling frameworks. Specifically, \textcolor{black}{the} SATE module first calculates the output distributions of known samples as an approximation of the distributions of unknown samples. It then estimates a suitable threshold for each category with the statistical properties of unknown samples, aiming to achieve a balance between precision and recall of pseudo-labels. Furthermore, the DRL is introduced to enhance the discriminative ability of the estimated thresholds. This loss function encourages the model to widen the gaps between positive and negative distributions using information from the known samples. Finally, through the incorporation of these complementary components, the model is able to obtain accurate pseudo-labels, which are used for model training along with the known labels. An overall illustration of our framework is shown in Figure \ref{Fig:SATL}.

We then introduce the notations we're going to use in this paper. We denote the training set as $D=\{(I^1, y^1), (I^2, y^2),...,(I^N, y^N)\}$, where $I^n$ is the $n$-th image and $y^n$ represents its label vector. In partial-label setting, only a small number of positive and negative labels are known while others are missing. For each sample $(I^n,y^n)$, the label vector can be written as $y^n=[ y^n_1, y^n_2, ..., y^n_C ] \in \{-1,0,1\}^C$, in which $C$ is the number of categories in $D$ and $y^n_c \in \{-1,0,1\}$ is the label of category $c$. $y^n_c$ is assigned to 1 if the category $c$ exists in $I^n$, assigned to -1 if it does not exist, and assigned to 0 if it is unknown.

\subsection{Pseudo-Labeling For Partial Label Learning}

In partial label learning, we attempt to gain extra supervision information given the incomplete annotations. Pseudo-labeling method is an ideal way to solve this problem by introducing new possible labels in the training process. Following prior works \cite{durand2019learning}, we show the main process of pseudo-labeling method in MLR-PL. 

Given an input image $I^n$, we first utilize a feature encoder to get its feature map. Then, a module is following to obtain an output distribution $\widetilde{p}^n=[\widetilde{p}^n_1, \widetilde{p}^n_2,...,\widetilde{p}^n_C]$. This module can either be the classification head of the model, or an extra pseudo-label generator. Using the output distributions, we assign samples with confidence scores higher than the threshold as positive and generate pseudo-labels, which can be formulated as follows.
\begin{equation} \label{eq2}
    \widetilde{y}^n_c=\textbf{1}[\widetilde{p}^n_c > \tau_c]
\end{equation}
where $\textbf{1}[\cdot]$ is an indicator function whose value is 1 if the argument is true and 0 otherwise. $\tau_c \in (0,1)$ is the confidence threshold for selecting pseudo-labels of $c$-th category. We can therefore use the pseudo-labels $\widetilde{y}^n=[\widetilde{y}^n_1,\widetilde{y}^n_2,...,\widetilde{y}^n_C]$ combining with the known labels for training. 

As illustrated above, the selection of $\tau_c$ in each category is crucial for high quality pseudo-label generation, which is precisely the focus of our methodology. 

\subsection{Semantic-Aware Threshold Estimation}

As illustrated above, a self-adaptive class-specific threshold adjustment scheme is crucial for model training. Since labeled and unlabeled objects from the same category exhibit semantic similarities, we can explicitly compute the output distributions of known samples, and employ them as reliable approximations of the unknown distributions. With the statistical information of unknown samples, we can estimate the ideal thresholds by considering the lower bound of error tolerance for pseudo-labels.

Motivated by above analysis, we first consider the output distributions of known samples $\{\widetilde{p}^n_c\}_{n=1}^{N_L}=\{ \widetilde{p}^1_c, \widetilde{p}^2_c,...,\widetilde{p}^{N_L}_c \}$, in which $N_L$ represents the number of labeled samples and $c$ denotes their respective categories. Then, we set two pre-defined parameters $\kappa^{-}, \kappa^{+}$ which serve as indicators of the uncertainty within the output distributions. With these two parameters in known distribution, two boundary line respective to positive and negative distributions are determined, which is formulated as Eq. (\ref{eq3}).
\begin{equation}\label{eq3}
    Prob[\widetilde{p}^n_c \leq \tau_{c}^{\pm}]=\kappa^{\pm}, \quad n=1,2,...,N_L
\end{equation}
These thresholds $\tau_{c}^{-}$ and $\tau_{c}^{+}$ are also applied to the distributions of unknown samples.

For a specific category, the threshold should be determined based on the statistical properties of the output distributions. The analysis of the distributions reveals two distinct situations. The first situation is that positive and negative samples are separated distinctly. In this situation, setting the threshold within the area of low density establishes a robust boundary to distinguish positive and negative samples, which allows model to recall more pseudo-labels under the premise of high accuracy. Conversely, in the second situation where the distributions of positive and negative samples are mixed up, the model could not confidently distinguish between positive and negative samples. Therefore, we set a comparatively high threshold to avoid false positive pseudo-labels. 

We can distinguish these two situations from the values of $\tau_c^{-}$ and $\tau_c^{+}$, and choose the threshold to achieve the balance between precision and recall. Therefore, we determine the ideal threshold $\tau_c^*$ for each category $c$ as Eq. (\ref{eq4}), based on the two distinct thresholds above:
\begin{equation}\label{eq4}
    \tau_c^*=max\{\tau_{c}^{-},\tau_{c}^{+} \}
\end{equation}

For each iteration $t$, we estimate the ideal threshold $\tau_c^*(t)$, and the model's threshold $\tau_c(t)$ is updated as shown in Eq. (\ref{eq5}),
\begin{equation}\label{eq5}
    \tau_c(t+1)= \gamma \tau_c(t) + (1-\gamma)\tau_c^*(t), \quad \gamma \in (0,1)
\end{equation}
where $\gamma$ is a hyper-parameter that controls the convergence rate of the thresholds. The selection of specific values for $\gamma$ will be further discussed in Section \ref{sec:analysis}.

\begin{figure}[H] 
	\centering 
	\includegraphics[width=0.9\linewidth]{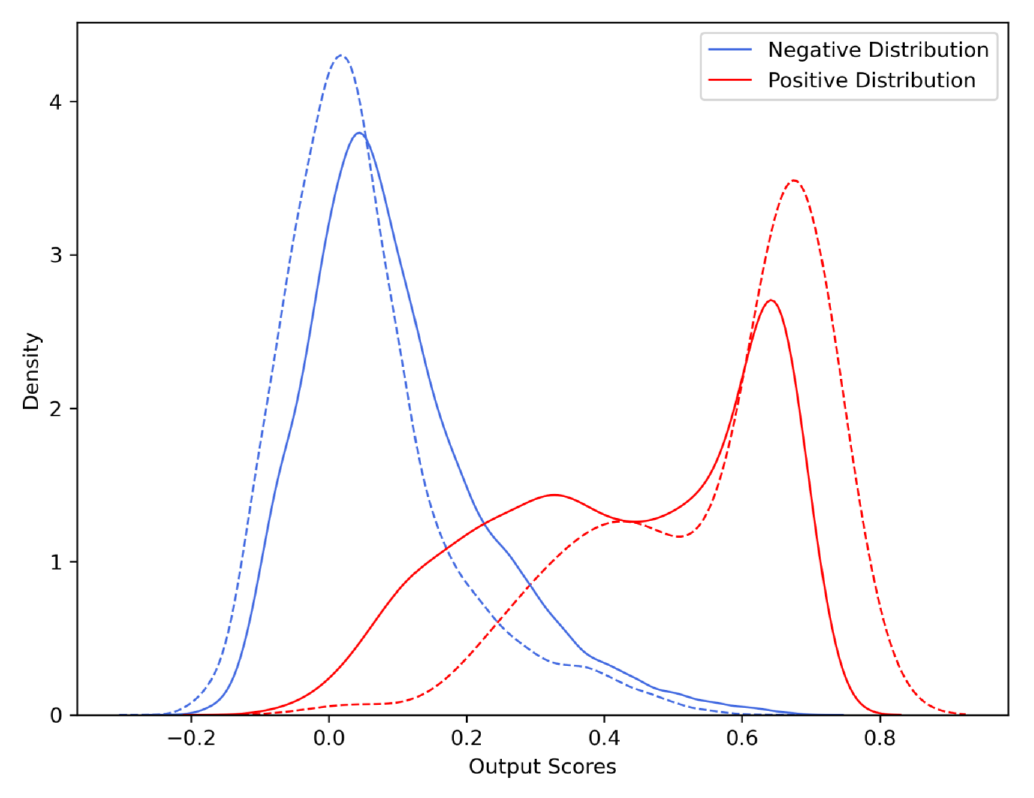}
	\caption{An illustration of when the positive and negative samples become more distinguishable as the gaps between their distributions widen (dashed lines) compared to the original distributions (solid lines). The DRL is proposed to enlarge the distances between positive and negative samples.}
	\label{Fig:DRL}
\end{figure}

\subsection{Differential Ranking Loss}

As illustrated above, the SATE module employs distinct methods to estimate class-specific thresholds based on the statistical distinction between positive and negative distributions within the category. When these distributions overlap significantly, the model encounters challenges in confidently discriminating between positive and negative samples. Consequently, it is forced to employ a conservative way in threshold estimation, leading to diminished recall of pseudo-labels. Thus, in addition to determining a reasonable threshold for each category, there arises a necessity to enlarge the distances between positive and negative samples. This enables the model to learn discriminative distributions and thereby improves both the recall and precision of pseudo-labels.

To achieve this goal, we propose a new differential ranking loss using the information of known samples. Given the produced distributions $\widetilde{p}^n=[\widetilde{p}^n_1, \widetilde{p}^n_2,...,\widetilde{p}^n_C]$, we consider its differences between the class-specific threshold vector $[\tau_1, \tau_2, ..., \tau_C]$ by
\begin{equation}\label{eq6}
    d^n_c= max (0, \widetilde{p}^n_c-\tau_c)
\end{equation}
In this way, we get the threshold difference vector $[d^n_1, d^n_2, ..., d^n_C]$, where $d^n_c \in [0,1]$ is the deviation between the confidence score and the learned threshold. We then proceed the supervision process with known labels. For a positive known label $y^n_c=1$ in image $n$, it is expected that the indicator function in Eq. \ref{eq2} is positive, so the difference $\widetilde{p}^n_c-\tau_c$ should also tend to be positive. Conversely, for a negative known label $y^n_c=-1$, the difference should be negative. Therefore, the positive and negative samples in each category are propelled in opposite directions, widening the gap between their distributions. From the above analysis, we consider this objective as a ranking task and formulate the differential ranking loss as
\begin{equation}\label{eq7}
    L_{drl}=\sum_{n=1}^{N} \sum_{c=1}^{C} s^{n,c}
\end{equation}
where $s^{n,c}$ is defined as follows: 
\begin{equation}\label{eq8}
    s^{n,c}=
    \begin{cases}
        1-d^{n,c}, & y^{n,c}=+1\\
        1+d^{n,c}, & y^{n,c}=-1
    \end{cases}
\end{equation}

During the training process, $d^n_c$ should tend to be large as $y^n_c$ is positive, and vice versa. As illustrated in Figure \ref{Fig:DRL}, the objective function encourages model to enlarge the gaps between positive and negative distributions, thereby enhancing the discriminative ability of the thresholds.

\subsection{Optimization}
Following previous works \cite{durand2019learning,chen2022structured}, we first use the partial BCE loss function as the basic objective function. Specifically, given the model prediction distribution $p^n=[p^n_1, p^n_2,...,p^n_C]$, we utilize the partial ground truth labels $y_n=[y^n_1, y^n_2,...,y^n_C]$ and define the objective function for the $n$-th sample as follows.
\begin{equation}\label{eq9}
    l(p^n,y^n)= \frac{1}{\sum_{c=1}^{C} |y^n_c|} \sum_{c=1}^{C} [\textbf{1}(y^n_c=1) log(p^n_c) + \textbf{1}(y^n_c=-1) log(1-p^n_c)]
\end{equation}

At the beginning of training process, the model is trained with partial labels to establish its basic capabilities. At the first stage of training, the objective function is defined as follows.

\begin{equation}\label{eq10}
    L_{cls}=\sum_{n=1}^N l(p^n,y^n)
\end{equation}

After sufficient training, pseudo-labels can be generated based on the class-specific thresholds mentioned above. We can then fuse the generated pseudo-labels $\widetilde{y}^n=[\widetilde{y}^n_1,\widetilde{y}^n_2,...,\widetilde{y}^n_C]$ with the known labels, and the result is denoted as $\hat{y}^n=[\hat{y}^n_1,\hat{y}^n_2,...,\hat{y}^n_C]$. The partial BCE loss is then modified as follows.
\begin{equation}\label{eq11}
    \hat{L}_{cls}=\sum_{n=1}^N l(p^n,\hat{y}^n)
\end{equation}
where $\hat{y}^n_c$ is defined by
\begin{equation}\label{eq12}
    \hat{y}^n_c=
    \begin{cases}
        y^n_c, & y^{n,c}=\pm 1\\
        \widetilde{y}^n_c, & y^{n,c}= 0
    \end{cases}
\end{equation}

As discussed above, we combine the differential ranking loss function with the proposed SATE module to refine the thresholds. Thus, in the second stage of model training, the overall loss function can be formulated as
\begin{equation}\label{eq13}
    L_{SATL}=\hat{L}_{cls} + \lambda L_{drl}
\end{equation}
Finally, based on the discussion above, we can summarize the entire process of SATL in Algorithm \ref{alg:Framwork}. \textcolor{black}{It is worth noting that during the model's inference stage, the pseudo-label generation module and threshold parameters are omitted. Only the trained model $f_{\tilde{\theta}}$ is used to perform inference on the image $I^n$ and generate the classification result $[p^n_1, p^n_2, ..., p^n_C]$.
}

\begin{algorithm}[!ht]
\caption{SATL algorithm.}
\label{alg:Framwork}
\begin{algorithmic}[1] 
    \REQUIRE $\alpha$, $\beta$: Learning rates
    \REQUIRE $C_{num}$: Number of category
    \REQUIRE $\theta$: Randomly initialized parameters of model \\
    \REQUIRE $D=\{(I^n,y^n)\}_{n=1}^{N}$: Dataset with partial labels, where $y^n \in \{ -1,0,1\}$ and N is the size of dataset\\
    \REQUIRE $E_{init}$, $E_{end}$: Epoch number for the first and second stage of training \\

    \WHILE {$epoch \le  E_{init}$}
        \STATE Make predictions $p^n$ using model with parameters $\theta$;
        \STATE Compute the value of loss function $L_{cls}$ in Eq \ref{eq10};
        \STATE Update parameters with gradient descent:\\ $\hat{\theta} \gets \theta-\alpha{\nabla_{\theta}}L_{cls}(\theta)$;
    \ENDWHILE

    \WHILE {$epoch \le  E_{end}$}
        \STATE Make predictions with parameters $\hat{\theta}$: $p^n=f_{\hat{\theta}}(I^n)$;
        \WHILE {$c \le C_{num}$}
            \STATE Compute the output distributions of known samples;
            \STATE Compute ideal threshold $\tau_c^*$ with Eq \ref{eq3} and Eq \ref{eq4};
            \STATE Update class-specific threshold $\tau_c$ with Eq \ref{eq5};
        \ENDWHILE
        \STATE Compute the value of loss function $L_{SATL}$ in Eq \ref{eq13};
        \STATE Update parameters $\hat{\theta}$ with gradient descent:\\ $\tilde{\theta} \gets \hat{\theta}-\beta{\nabla_{\hat{\theta}}}L_{SATL}(\hat{\theta})$;
    \ENDWHILE
    \ENSURE ~~\\ 
        Updated model parameter $\tilde{\theta}$.
\end{algorithmic}
\end{algorithm}

\section{Experiments}
In this section, we conduct detailed comparison on various datasets and method analysis to have a better understanding.

\subsection{Experimental Settings}
We first illustrate the dataset used for model comparison and analysis. Following previous works, we implement our experiments on the MS-COCO \cite{lin2014microsoft} and Visual Genome \cite{krishna2017visual} dataset. \textcolor{black}{We adopt these two datasets because they cover a wide range of commonly used categories and diverse scenarios in practical applications. For the MLR-PL task, this enables the model to address the problem of partial labels across various scenarios, thereby demonstrating the generalizability and applicability of the proposed method}. The MS-COCO dataset contains about 120k images and 80 different categories in daily life. This dataset is further divided into a training set with approximately 80k images and a validation set with approximately 40k images. The Visual Genome is a large dataset with visual instances and their text descriptions, containing 108249 images and 80138 different categories. Since most of these categories are contained only in few samples, we consider the 200 most-frequent categories and construct the VG-200 dataset. Then, we randomly choose 10000 images as validation set and the rest 98249 images as training set.

Since these datasets are fully annotated, while our work focus on partially annotated datasets, we follow the previous work and randomly drop part of the labels to create a partially annotated training set. \textcolor{black}{Specifically, for each sample, we generate a random vector of length equal to the number of categories, where each component is independently sampled from a uniform distribution over $[0,1]$. Labels corresponding to components below the specified label proportion are retained, while the label information for the remaining positions is discarded. In this way, we construct a dataset with a predefined known label proportion without introducing bias
}. In our experiments, we consider the scenarios with very low label availability, where the proportion of unknown labels ranges from 95\% to 50\%, resulting in only 5\% to 50\% of the labels are known to the model. 

\begin{figure}[b] 
	\centering 
	\includegraphics[width=0.9\linewidth]{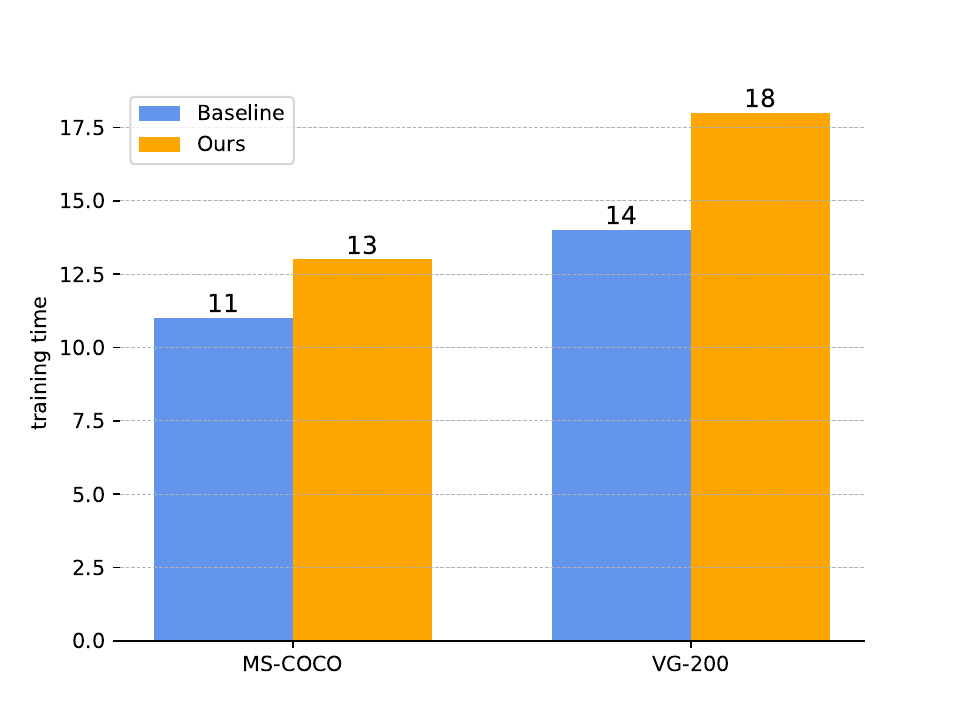}
	\caption{\textcolor{black}{Comparison of training time of baseline (blue) and baseline with SATL module (orange) in different datasets.}}
	\label{Fig:trainingTime}
\end{figure}

\begin{table*}[!ht] 
    \centering 
    \caption{\textcolor{black}{The average mAP and mAP values of our SATL framework and current state-of-the-art methods for MLR-PL with different known label proportions on the MS-COCO and VG-200 datasets. Our results are highlighted in bold. The proposed method outperforms the baseline methods under all known label proportions.}} 
    \vspace{5pt} 
    \begin{tabular*}{\textwidth}{@{\extracolsep{\fill}} ccccccccc}
    \hline 
    Datasets &Methods &5\% &10\% &20\% &30\% &40\% &50\% &Avg.mAP\\ 
    \hline
    \multirow{7}{*}{\centering MS-COCO}
    &SSGRL  &$62.2\pm0.4$ &$67.6\pm0.4$ &$70.5\pm0.4$ &$72.1\pm0.3$ &$73.4\pm0.2$ &$74.8\pm0.2$ &$69.9\pm0.3$\\
    &GCN-ML  &$62.1\pm0.2$ &$66.4\pm0.2$ &$70.2\pm0.2$ &$72.4\pm0.2$ &$73.6\pm0.1$ &$74.9\pm0.1$ &$69.9\pm0.2$\\
    &KGGR  &$66.3\pm0.2$ &$70.9\pm0.1$ &$73.1\pm0.2$ &$75.4\pm0.1$ &$76.7\pm0.1$ &$77.5\pm0.1$ &$73.3\pm0.1$\\
    &SST  &$65.8\pm0.2$ &$68.1\pm0.2$ &$73.5\pm0.1$ &$75.9\pm0.2$ &$77.3\pm0.2$ &$78.1\pm0.1$ &$73.1\pm0.2$\\
    &HST  &$66.3\pm0.4$ &$70.6\pm0.3$ &$75.8\pm0.3$ &$77.3\pm0.3$ &$78.3\pm0.2$ &$79.0\pm0.2$ &$74.5\pm0.3$\\
    &\textbf{SST+SATL} &\textbf{68.9$\pm$0.3} &\textbf{73.7 $\pm$ 0.3} &\textbf{76.6 $\pm$ 0.2} &\textbf{78.0$\pm$0.2} &\textbf{78.9$\pm$0.2} &\textbf{80.1$\pm$0.1} &\textbf{76.0$\pm$0.2}\\
    &\textbf{HST+SATL} &\textbf{69.1$\pm$0.4} &\textbf{74.0$\pm$0.3} &\textbf{77.5$\pm$0.3} &\textbf{78.5$\pm$0.3} &\textbf{79.5$\pm$0.2} &\textbf{80.3$\pm$0.2} &\textbf{76.5$\pm$0.3}\\
    \hline
    \multirow{7}{*}{\centering VG-200}
    &SSGRL  &$31.7\pm0.9$ &$35.1\pm0.7$ &$38.0\pm0.6$ &$38.9\pm0.4$ &$39.9\pm0.4$ &$40.4\pm0.3$ &$37.3\pm0.5$\\
    &GCN-ML  &$29.3\pm0.6$ &$33.2\pm0.4$ &$38.1\pm0.3$ &$39.2\pm0.3$ &$40.1\pm0.3$ &$41.3\pm0.2$ &$36.9\pm0.3$\\
    &KGGR  &$35.8\pm0.4$ &$38.6\pm0.3$ &$40.1\pm0.2$ &$41.3\pm0.2$ &$42.5\pm0.1$ &$43.3\pm0.1$ &$40.3\pm0.2$\\
    &SST  &$36.5\pm0.4$ &$38.8\pm0.3$ &$39.4\pm0.2$ &$41.1\pm0.2$ &$41.8\pm0.1$ &$42.7\pm0.1$ &$40.1\pm0.2$\\
    &HST  &$35.0\pm0.6$ &$38.7\pm0.5$ &$41.5\pm0.3$ &$43.3\pm0.3$ &$44.6\pm0.3$ &$45.2\pm0.2$ &$42.6\pm0.3$\\
    &\textbf{SST+SATL} &\textbf{39.7$\pm$0.4} &\textbf{43.1$\pm$0.3} &\textbf{45.9$\pm$0.3} &\textbf{47.0$\pm$0.3} &\textbf{47.7$\pm$0.2} &\textbf{48.2$\pm$0.2} &\textbf{45.3$\pm$0.3}\\
    &\textbf{HST+SATL} &\textbf{39.8$\pm$0.5} &\textbf{43.0$\pm$0.5} &\textbf{45.2$\pm$0.4} &\textbf{47.0$\pm$0.3} &\textbf{47.7$\pm$0.3} &\textbf{48.3$\pm$0.2} &\textbf{45.3$\pm$0.3}\\
    \hline
\end{tabular*}
\label{table1} 
\end{table*}

\subsection{Evaluation metrics}

For fair comparison, we follow previous works \cite{durand2019learning, chen2022structured} and adapt the mean average precision (mAP) over all categories under different proportions of known labels as above. The label proportions are set to 5\%, 10\%, 20\%, 30\%, 40\% and 50\%. To visually compare the performance of different methods, we compute the average mAP over all proportion settings. Moreover, we use other standard multi-label classification metrics for more comprehensive comparisons with other methods, including the overall and per-class F1-measure (i.e., OF1 and CF1). The OF1 and CF1 scores can be computed by
\begin{equation}
    OP=\frac{\sum_i N^c_i}{\sum_i N^p_i}, \quad CP=\frac{1}{C} \frac{\sum_i N^c_i}{\sum_i N^p_i}
\end{equation}
\begin{equation}
    OR=\frac{\sum_i N^c_i}{\sum_i N^g_i}, \quad CR=\frac{1}{C} \frac{\sum_i N^c_i}{\sum_i N^g_i}
\end{equation}
\begin{equation}
    OF1=\frac{2 \times OP \times OR}{OP+OR}, \quad CF1=\frac{2 \times CP \times CR}{CP+CR}
\end{equation}
where $C$ is the number of labels, $N^c_i$ is the number of images that are correctly predicted in the $i$-th category, $N^p_i$ is the number of predicted in the $i$-th category, and $N^g_i$ is the number of ground truth images in the $i$-th category. Among these metrics, mAP, OF1, CF1 are the most important metrics for comprehensive comparisons, and we mainly use them as our evaluation metrics.

\subsection{Baseline and Implementation Details}

As is shown above, the proposed SATL framework can be used as an extra component for pseudo-labeling method in MLR-PL. We choose the representative SST \cite{chen2022structured} and HST \cite{chen2022heterogeneous} frameworks as our baseline methods, since they both introduce semantic transfer modules to obtain pseudo-labels, which are suitable to our proposed framework. All the settings of network structures and hyperparameters keep aligned with the original SST and HST methods. The training objective of the whole framework is modified according to Eq. (\ref{eq13}). At the second stage of training, $\lambda_{IST}, \lambda_{CST}$ for IST and CST are both set to 0.01. The uncertainty values for both module $\kappa^{-}, \kappa^{+}$ are set to 0.999, 0.1 respectively in order to filter out pseudo-labels in a rather high precision. The momentum coefficients for threshold updating $\gamma_{IST}, \gamma_{CST}$ for IST and CST module are set to the same value according to the proportion of known labels, which will be further discussed in Section \ref{sec:analysis}. During the training process, the initial threshold is set to 1.0 for the first stage to avoid misleading pseudo-labels. 

\textcolor{black}{Our model is trained on a single 24GB NVIDIA RTX 3090 GPU. A comparison of the training times between the baseline method and the model enhanced with the proposed SATL module is presented in Figure \ref{Fig:trainingTime}. The training time for the SATL-enhanced model increases by 2 hours and 4 hours, respectively. This increase is primarily due to the computation of histograms for the output distribution and the calculation of the thresholds $\tau^+_c$ and $\tau^-_c$. Moreover, the proposed SATL module introduces only additional parameters for category-specific thresholds, without significantly increasing the model's complexity or training cost}.

\subsection{Comparison With Baseline Methods}

To evaluate the effectiveness of the proposed SATL framework, we first implement it to the SST \cite{chen2022structured} and HST \cite{chen2022heterogeneous} frameworks as stated above. Then, we compare our methods with the following algorithms, which can be divided into two groups. (1) The SSGRL \cite{chen2019learning}, ML-GCN \cite{chen2019multi} and KGGR \cite{chen2022knowledge}. These methods introduce graph neural networks to model label dependencies, and they achieve state-of-the-art performance on the traditional MLR task. To adapt these methods to address MLR-PL task, we replace the BCE loss with the partial BCE loss as shown in Eq. (\ref{eq9}) while keeping other parts unchanged. (2) The baseline methods, namely the SST framework \cite{chen2022structured} and HST framework \cite{chen2022heterogeneous}. Since they are suitable for the task of MLR-PL, we follow all the settings detailed in these works.

\begin{table}[!t]
    \centering
    \caption{The average OF1 and CF1 values achieved by our SATL framework compared to current state-of-the-art methods for MLR-PL on the MS-COCO and VG-200 datasets. The proposed method achieves superior performance compared to the current mainstream methods.}
    \vspace{5pt}
    \begin{tabular}{c|c|c|c}
    \hline
        Datasets & Methods &Avg.OF1 &Avg.CF1\\
        \hline
        \multirow{7}{*}{\centering MS-COCO}
         & SSGRL &71.6 &64.6  \\
         & GCN-ML &70.9 &64.7  \\
         & KGGR &73.8 &69.0 \\
         & SST &72.9 &68.7 \\
         & HST &73.7 &69.7 \\
         & \textbf{SST+SATL} &\textbf{75.4} &\textbf{70.0} \\
         & \textbf{HST+SATL} &\textbf{76.0} &\textbf{71.5} \\
        \hline
        \multirow{7}{*}{\centering VG-200}
         & SSGRL &35.3 &23.4  \\
         & GCN-ML &36.9 &23.9  \\
         & KGGR &40.0 &32.3 \\
         & SST &41.2 &32.0 \\
         & HST &40.6 &31.5 \\
         & \textbf{SST+SATL} &\textbf{46.9} &\textbf{40.4} \\
         & \textbf{HST+SATL} &\textbf{46.0} &\textbf{39.5} \\
        \hline
    \end{tabular}
    \label{table2}
\end{table}

\begin{table}[ht]
    \small
    \centering
    \caption{\textcolor{black}{The t-values and p-values for paired \textbf{t}-test between the variants of proposed method (SST+SATL and HST+SATL) and comparative methods, with significant level $\alpha=0.05$.}}
    \vspace{5pt}
    \begin{tabular}{c|c|c|c}
    \hline
        Proposed methods & Comparative methods &t-values &p-values\\
        \hline
        \multirow{7}{*}{\centering SST+SATL}
         & SSGRL &7.0953 &0.0008  \\
         & GCN-ML &7.2766 &0.0006  \\
         & KGGR &6.6764 &0.0008 \\
         & SST &4.8824 &0.0019 \\
         & HST &3.7832 &0.0150 \\
        \hline
        \multirow{7}{*}{\centering HST+SATL}
         & SSGRL &7.6657 &0.0001  \\
         & GCN-ML &7.8352 &0.0001  \\
         & KGGR &7.2193 &0.0006 \\
         & SST &4.9549 &0.0021 \\
         & HST &3.9761 &0.0132 \\
        \hline
    \end{tabular}
    \label{table_ttest}
\end{table}

\subsubsection{Performance on MS-COCO}
MS-COCO is the most commonly used dataset in MLR evaluation. We present the comparison results of mAP under different label proportions and the average OF1, CF1 scores to provide more comprehensive evaluations, as shown in Table \ref{table1} and \ref{table2}. Comparing with the SST framework, our proposed framework achieves average mAP, OF1, CF1 values of 76.0\%, 75.4\%, 70.0\%, with average improvements of 2.9\%, 2.5\%, 1.3\%, respectively. Also, compared to the HST framework, our proposed framework achieves average mAP, OF1, CF1 values of 76.5\%, 76.0\%, 71.5\%, with average improvements of 2.0\%, 2.3\%, and 1.8\%. In addition to these average results, our method also achieves better results under all label proportion settings, especially in the cases of very low proportions. These results show that equipped with the proposed SATL framework, the MLR-PL methods exhibit improvement since more pseudo-labels are recalled and utilized in the training process. Based on the SST and HST framework, our method also outperform other mainstream fully-label approaches across all known label proportion settings. KGGR adapts graph semantic propagation to update its classifier, achieving rather high performance among all label proportions. However, it remains less competitive compared to our method due to the lack of supervision information from pseudo-labels.

\subsubsection{Performance on VG-200}
VG-200 dataset is a more challenging dataset compared to the MS-COCO dataset, since it covers more categories. Similar to the MS-COCO dataset, we use the mAP, OF1 and CF1 scores as our metrics and present the comparison results on VG-200 dataset as shown in Table \ref{table1} and \ref{table2}. Comparing with the SST framework, we obtain improvements in average mAP, OF1, CF1 with 5.2\%, 5.7\% and 8.2\% respectively, since more high quality pseudo-labels are provided during the training process. Meanwhile, compared to the HST framework, our framework achieves the best average mAP, OF1 and CF1 with 45.3\%, 46.0\% and 39.5\%, with improvement of 2.7\%, 5.4\% and 8.0\%. We also present the mAP results under all label proportion settings, as shown in Table \ref{table1}. These results demonstrate a more obvious improvement compared to those of the MS-COCO dataset, which confirms that the proposed SATL method remains robust even as the number of categories and unknown labels increase. 

\subsubsection{\textcolor{black}{Statistical test}}
\textcolor{black}{To validate the statistical significance of our method's performance improvement over existing approaches, we conduct paired \textbf{t}-tests between the proposed method and all comparative methods. The \textbf{t}-value for the paired \textbf{t}-test is defined as follows:
\begin{equation}
    t = \frac{\bar{x}_d}{s_d/ \sqrt{n}},
\end{equation}
where $\bar{x}_d$ denotes the mean of paired differences, $s_d$ represents the standard deviation of paired differences, and $n$ is the number of paired samples. We perform all tests using a two-tailed setting and set the significance level at $\alpha = 0.05$. 
}

\textcolor{black}{
In our experimental design, mAP scores are treated as paired measurements between each proposed variant (i.e., the SST+SATL and the HST+SATL) and comparative method.
Meanwhile, all models are trained on the MS-COCO dataset under identical random seeds, with experiments repeated over multiple independent choices of seed to generate paired samples. The results of the statistical test are shown in Table \ref{table_ttest}. These results demonstrate p-values $<$ 0.05 for all comparisons, which are substantially below the $\alpha$ threshold. This provides sufficient evidence to reject the null hypothesis $H_0$ (which states no significant difference between the proposed and comparative methods). Consequently, we conclude that the proposed method achieves statistically superior mean average precision compared to current leading methods.
}

\begin{figure*}[!h] 
	\centering 
	\includegraphics[width=1.0\linewidth]{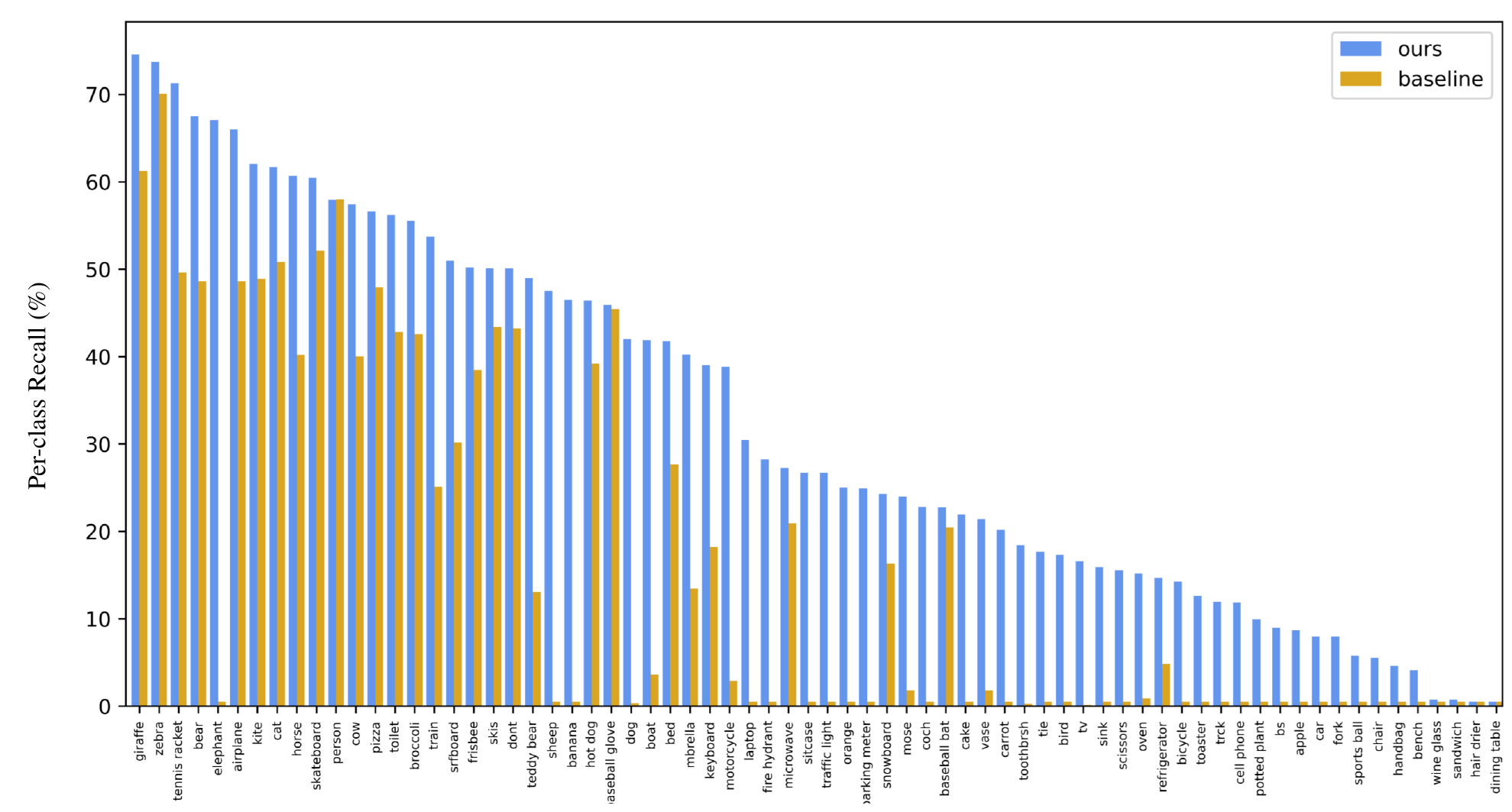}
	\caption{Recall rate in pseudo-labels under the threshold settings of SATL and SST, under the proportion of 20\% known labels. Our method recalls more pseudo-labels in most categories of the MS-COCO dataset.}
	\label{Fig:recall}
\end{figure*}

\begin{figure}[t]
	\centering 
	\includegraphics[width=1.0\linewidth]{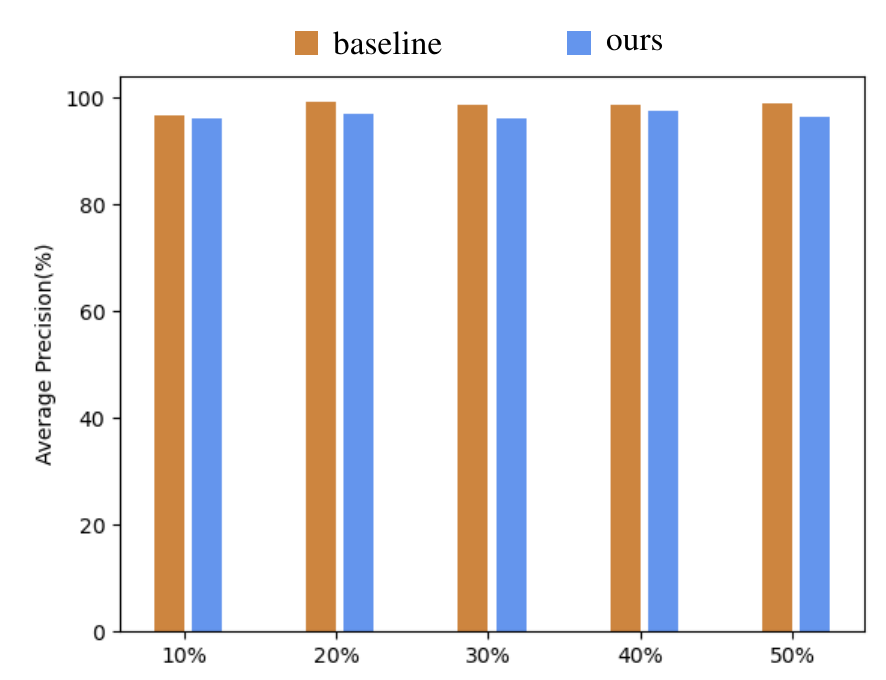}
	\caption{The average precision of pseudo-labels in unknown labels across all settings of the proportion. Our method maintains high-quality pseudo-labels, even when the quantity of such labels is increased.}
	\label{Fig:precision}
\end{figure}

\begin{figure*}[htbp] 
	\centering 
	\includegraphics[width=1.0\linewidth]{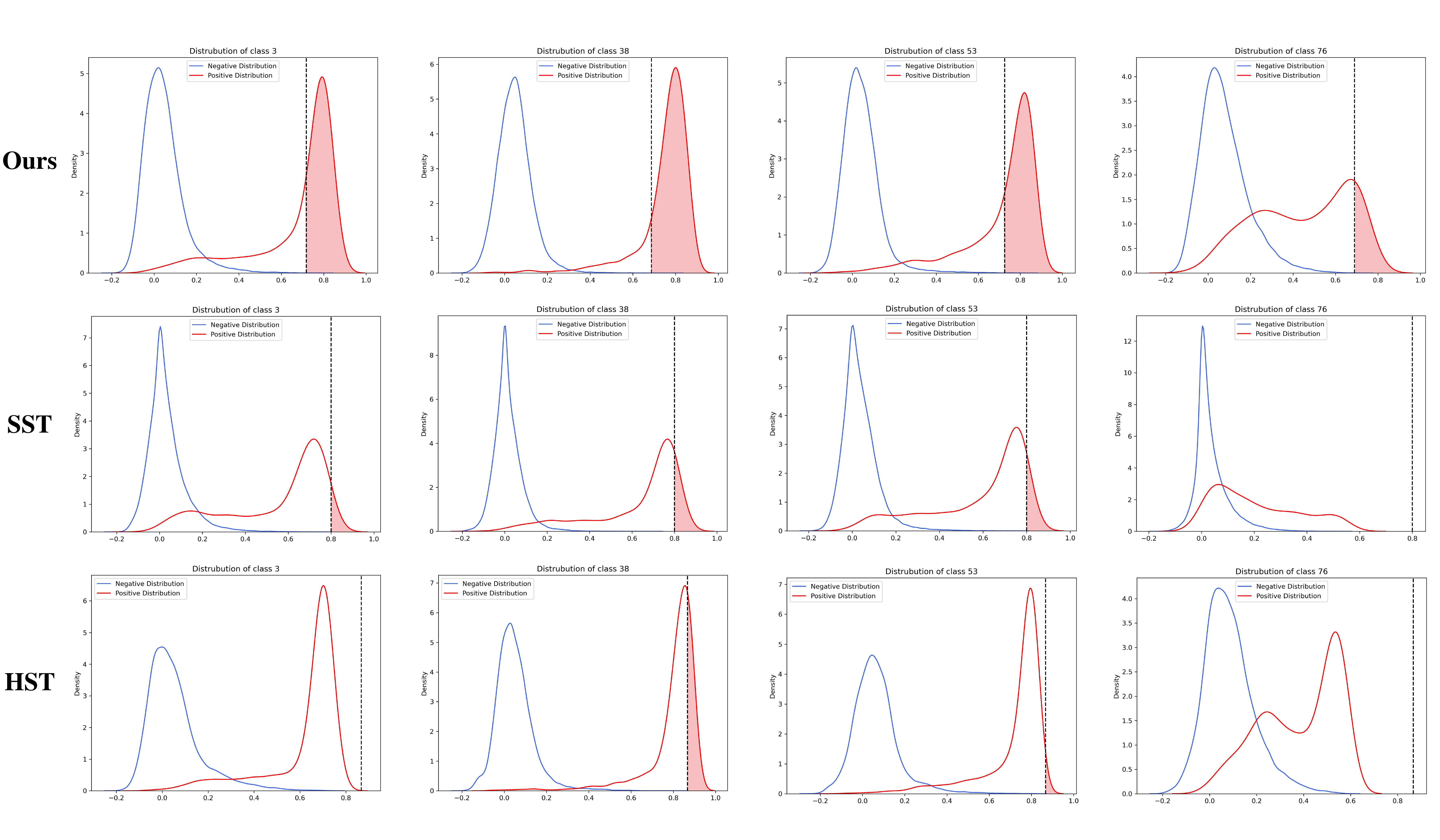}
	\caption{Visualization of class-specific thresholds and the distributions of model output scores. The top row represents the distributions of the SATL framework, while the subsequent two rows represent the baseline methods. The labels being recalled are highlighted in red.}
	\label{Fig:visDiffMethod}
\end{figure*}

\begin{figure}[!h] 
	\centering 
	\includegraphics[width=1.0\linewidth]{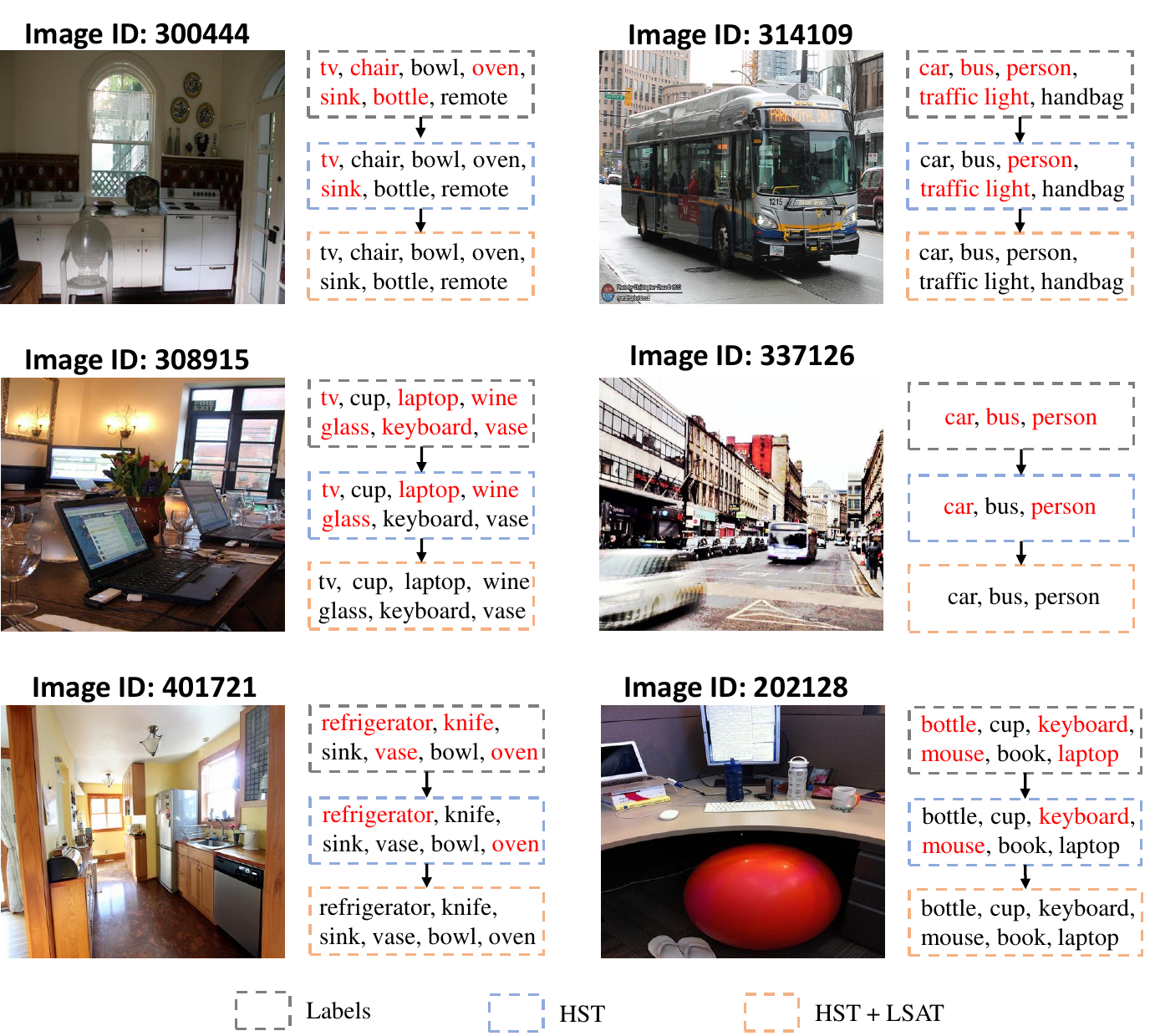}
	\caption{A comparative illustration of the recalled labels from HST and our method under the label proportion of 20\%. The unknown labels are indicated in red and the accessible labels (including both known and recalled labels) are in black. These pictures are randomly sampled from the MS-COCO dataset.}
	\label{Fig:visExp}
\end{figure}

\subsection{Method Analysis}

\subsubsection{Analysis of Pseudo-Labels}

To have a comprehensive understanding of SATL method, we further perform a detailed analysis of pseudo-label quality and the visualization of class-specific thresholds. Specifically, we conduct experiments under the setting of 20\% known labels on MS-COCO dataset. 

We first consider the recall and precision ratio of pseudo-labels in unknown labels. For each category $c$, we only consider unknown samples and calculate the per-class precision $P_c$ and recall $R_c$ of pseudo-labels with true positive ($TP_c$), false positive ($FP_c$) and false negative ($FN_c$).
\begin{equation}\label{eq17}
    P_c=\frac{TP_c}{TP_c+FP_c}, \quad R_c=\frac{TP_c}{TP_c+FN_c}
\end{equation}
As illustrated in Figure \ref{Fig:recall}, our SATL method outperforms the baseline SST method as more missing labels are recalled. Also, to examine the quality of generated pseudo-labels, we also visualize the average precision of pseudo-labels over unknown labels under different label proportions, as presented in Figure \ref{Fig:precision}. We can see that the quality of pseudo-labels are high enough to provide accurate supervision information for model's training.

To further explain the results mentioned above, we visualize the positions of thresholds in the model's output distributions by category. We compare the threshold positions from the SST, HST, and our method, as shown in Figure \ref{Fig:visDiffMethod}, and visualize the distributions from four different categories at epoch 10. The positive and negative distributions are distinguished with different colors. We observe that our class-specific thresholds lead to a higher recall of positive labels, whereas the SST and HST frameworks apply relatively higher thresholds and miss more true positive labels. Meanwhile, as illustrated before, an ideal threshold for pseudo-labels should rely on the low-density area between positive and negative distributions when these distributions are distinctly separate, thus enabling the selection of more missing labels. Otherwise, when the positive and negative distributions significantly overlap, the threshold should be adjusted to be high enough to prevent the inclusion of false positive labels. The illustration demonstrates that the thresholds estimated by our SATL method meet these criteria. 

To better understand the effect of the proposed SATL module, we present visual examples from the MS-COCO dataset, comparing pseudo-labels generated by the baseline method with those from our approach. As shown in Figure \ref{Fig:visExp}, with more reasonable thresholds, our method recalls more labels even for objects that are relatively small or vague. This suggests that while the baseline method tends to apply a uniformly high threshold for all categories, our method adapts class-specific thresholds to recall more labels while maintaining high accuracy.


\subsubsection{Analysis of the SATE module}
In our framework, the SATE module is proposed to estimate an ideal threshold for each category, which is further utilized for the iteration of class-specific thresholds. To demonstrate the effectiveness of SATE, we conduct component analysis under all settings of label proportions. We set SST as our baseline method, and add the SATE module individually to see the comparison results. As presented in Table \ref{table3}, the baseline SST method obtains the average mAP of 73.1\%, 40.1\% on the MS-COCO and VG-200 datasets. By introducing the SATE to estimate more reasonable thresholds, it improves average mAP to 75.6\%, 44.8\% on these datasets, with the improvement of 2.5\%, 4.7\%, respectively. These results demonstrate that our proposed threshold estimation method is helpful for complementing unknown labels and introducing more supervisory information that benefits model's training process.

\subsubsection{Analysis of the DRL}
The DRL is introduced to widen the gaps between positive and negative distributions in each category, thus improving the discriminative ability of the thresholds. Similar to the SATE module, we also conduct component analysis based on the SST baseline under all settings of label proportions. As illustrated in Table \ref{table3}, employing only the DRL ('ours DRL') yields the average mAP of 75.3\%, 44.2\% on MS-COCO and VG-200 datasets, with the improvement of 2.2\%, 4.1\%, respectively. Moreover, when the DRL is additionally incorporated, compared to the results of 'ours SATE', the model achieves further improvement of 0.4\%, 0.5\% on MS-COCO and VG-200 datasets. This illustrates the complementary relationship between the DRL and SATE modules, enabling the model to achieve better performance with the estimation of more reasonable thresholds.

\begin{table*}[!t] 
    \centering 
    \caption{Ablation study on MS-COCO and VG-200 datasets with different proportions of known labels. SST, Ours DRL and Ours SATE respectively represent the baseline SST method (with linear decay threshold), SST with differential ranking loss, and SST with semantic-aware threshold estimation method. And the Ours represents SST with both SATE and DRL.}
    \vspace{5pt} 
    \begin{tabular*}{\textwidth}{@{\extracolsep{\fill}} ccccccccc}
    \hline 
    Datasets &Methods &5\% &10\% &20\% &30\% &40\% &50\% &Avg.mAP\\ 
    \hline
    \multirow{4}{*}{\centering MS-COCO}
    &SST  &65.8 &68.1 &73.5 &75.9 &77.3 &78.1 &73.1\\
    &Ours SATE  &68.2 &73.5 &76.1 &77.9 &78.6 &79.5 &75.6\\
    &Ours DRL  &67.8 &73.4 &76.1 &77.4 &78.2 &78.7 &75.3\\
    &Ours  &\textbf{68.9} &\textbf{73.7} &\textbf{76.6} &\textbf{78.0} &\textbf{78.9} &\textbf{80.1} &\textbf{76.0}\\
    \hline
    \multirow{4}{*}{\centering VG-200}
    &SST  &36.5 &38.8 &39.4 &41.1 &41.8 &42.7 &40.1\\
    &Ours SATE  &39.2 &42.9 &45.8 &46.4 &47.0 &47.4 &44.8\\
    &Ours DRL  &38.3 &42.2 &45.1 &46.0 &46.7 &47.3 &44.2\\
    &Ours  &\textbf{39.7} &\textbf{43.1} &\textbf{45.9} &\textbf{47.0} &\textbf{47.7} &\textbf{48.2} &\textbf{45.3}\\
    \hline
\end{tabular*}
\label{table3} 
\end{table*}

\subsubsection{Analysis of Hyper-parameters}\label{sec:analysis}
As is discussed above, the threshold updating coefficient $\gamma$ is a crucial item that controls the convergence rate to the estimated thresholds. Considering that model's estimation of the appropriate thresholds may not be accurate in the early stages of training with only partial labels, it is necessary to control the threshold convergence rate. At the early stage of training, setting $\gamma$ to a small value may lead to conservative prediction of pseudo-labels, while setting $\gamma$ to a large value may recall some false positive pseudo-labels. Thus, this coefficient directly relates to the precision and recall of pseudo-labels. 

\begin{figure}[!t]
	\centering 
	\includegraphics[width=1.0\linewidth]{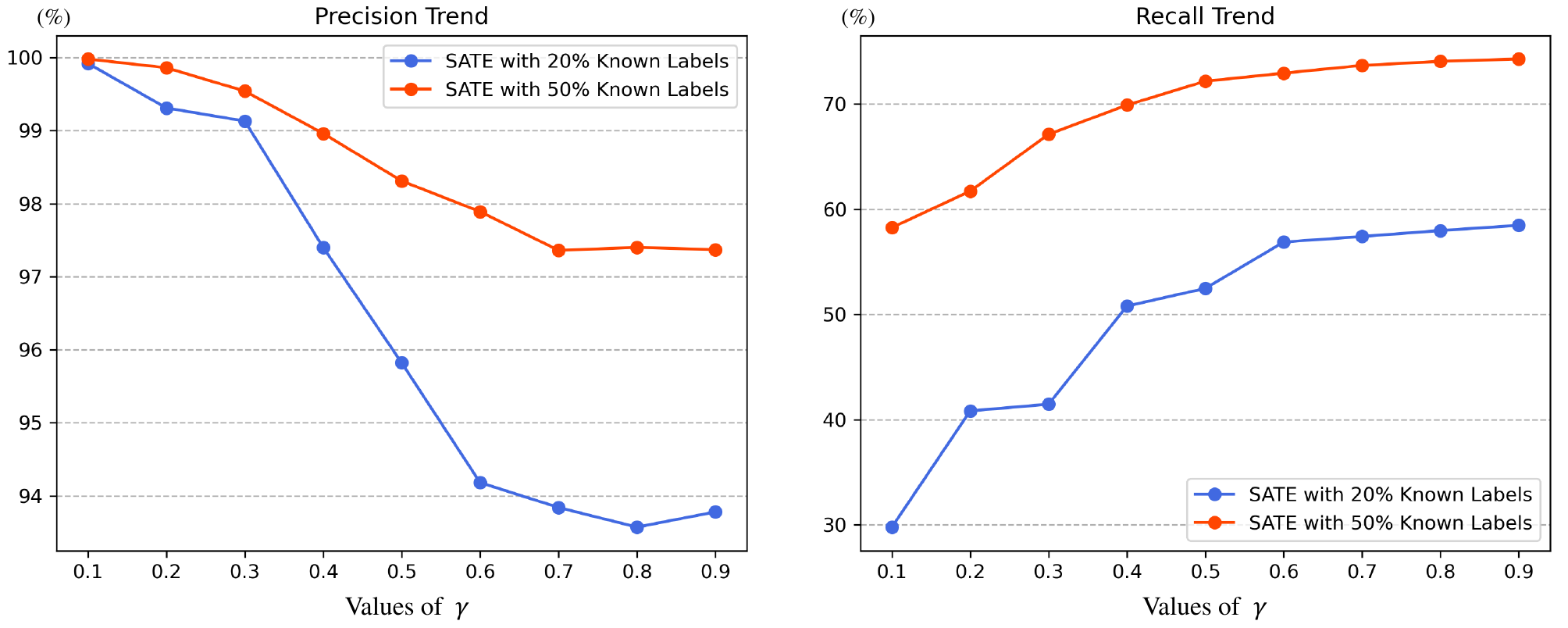}
	\caption{Analysis of the precision and recall trend of pseudo-labels under different hyper-parameter $\gamma$ on MS-COCO dataset with 20\% (blue lines) and 50\% (red lines) known labels. Only the SATE module is used in these experiments.}
	\label{Fig:curves}
\end{figure}

In order to study the settings of $\gamma$ under different label proportions, we conduct experiments on HST framework with $\gamma$ varying from 0.1 to 0.9 under the settings of 20\% and 50\% known labels on MS-COCO dataset, and we omit the use of DRL. We choose the precision and recall rates when the model achieves the optimal epoch performance respectively, and observe the corresponding label accuracy and recall curves, as shown in Figure \ref{Fig:curves}. 
It is observed that as $\gamma$ increases, the precision of pseudo-labels gradually decreases. Specifically, at label proportion of 20\%, the decline in the precision of pseudo-labels occurs more rapidly with the increase in $\gamma$ compared to the scenario with 50\%. This leads to more inaccurate pseudo-labels and degrades the model's performance, while the thresholds estimated with 50\% known labels are more robust. Meanwhile, as $\gamma$ increases, the growth rate of pseudo-label recall gradually decelerates. These observations indicate a need to balance the precision and recall of pseudo-labels. When the known label proportion is low, it is suitable to set $\gamma$ to a smaller value, whereas for a higher label proportion, setting $\gamma$ to a larger value is more appropriate. It is also crucial to choose the value before the trend of recall reaches a plateau. In the cases with 20\% and 50\% known labels, we set $\gamma$ to 0.3 and 0.5, respectively.

\begin{figure}[!h]
	\centering 
	\includegraphics[width=1.0\linewidth]{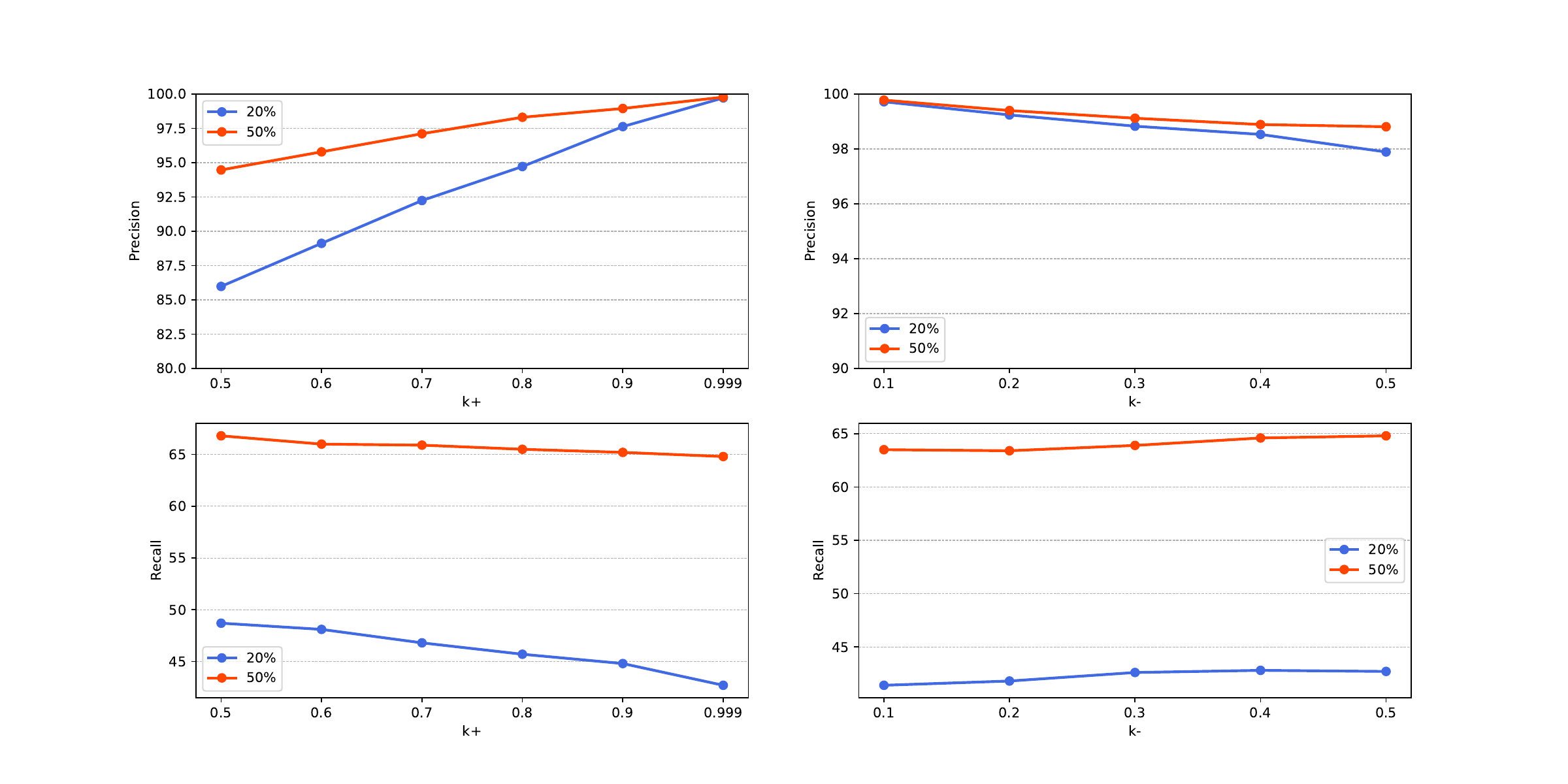}
	\caption{\textcolor{black}{Analysis of the precision (upper part) and recall (lower part) for pseudo-labels on the MS-COCO dataset with 20\% (blue) and 50\% (red) known labels with the variation of $\kappa^+$ and $\kappa^-$. Only the SATE module is used in these experiments.}}
	\label{Fig:kappa}
\end{figure}

\textcolor{black}{Moreover, $\kappa^+$ and $\kappa^-$ are crucial parameters that determine the category thresholds. They define the uncertainty tolerance for the positive and negative distributions, and they ensure that the estimated thresholds generate high-precision pseudo-labels. To study the impact of different values for $\kappa^+$ and $\kappa^-$, we conduct experiments on the MS-COCO dataset with 20\% and 50\% known labels, with SST as our main model. As in previous experiments, we omit the use of DRL in order to directly observe the effects of $\kappa^+$ and $\kappa^-$ on the category thresholds. As shown in Figure \ref{Fig:kappa}, we present the variation in the precision and recall of pseudo-labels as $\kappa^+$ changes from 0.999 to 0.5, with $\kappa^-$ fixed at 0.1. Under the 20\% label proportion, the precision of the pseudo-labels drops from 99.7\% to 85.9\%, while the recall shows a smaller increase from 34.7\% to 42.7\%. Furthermore, under the 50\% label proportion, the precision and recall of the pseudo-labels exhibit relatively less sensitivity to changes in $\kappa^+$, with variations of 5.3\% and 2.0\%, respectively. We also explore the impact of $\kappa^-$ as it changes from 0.1 to 0.5, while keeping $\kappa^+$ fixed at 0.999. In the scenarios with 20\% and 50\% known labels, the precision of the pseudo-labels increases by 1.8\% and 0.9\%, respectively, while recall decreases by 1.2\% and 1.3\%. When the label proportion is low, the prediction uncertainty is higher. To preserve high-quality pseudo-labels without affecting the stability of model training, we set $\kappa^+$ to a relatively high value. For higher label proportions, $\kappa^+$ can be slightly reduced to achieve higher pseudo-label recall. Meanwhile, the model is less sensitive to variations in $\kappa^-$ under different  label proportions. Nevertheless, to ensure the applicability across all categories, $\kappa^-$ should be set a low value.
}

\begin{figure}[!t]
	\centering 
	\includegraphics[width=1.0\linewidth]{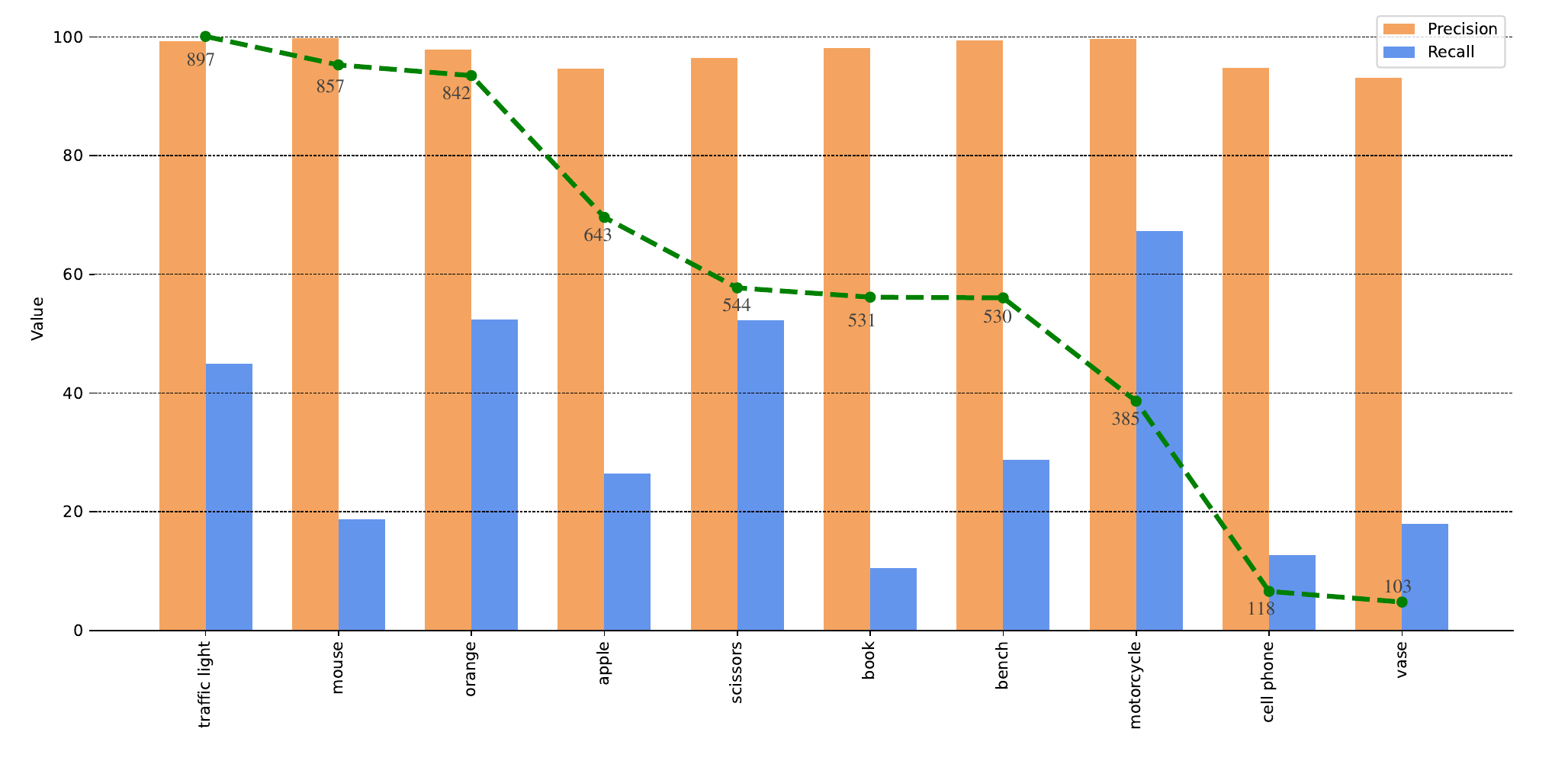}
	\caption{\textcolor{black}{Analysis of the precision (orange) and recall (blue) of pseudo-labels for the 10 least-represented categories in the MS-COCO dataset, with 20\% known labels. The number of samples in these categories is marked on the green line. These experiments are implemented with SST as the main model.}}
	\label{Fig:least10}
\end{figure}

\subsubsection{\textcolor{black}{Analysis of Data Imbalance Effects}}
\textcolor{black}{Due to the class imbalance inherent in real-world data, some categories contain fewer samples, which can lead to higher uncertainty in the output distribution, thereby influencing the determination of category thresholds. This long-tail effect is also observed in the MS-COCO and VG-200 datasets. To study the impact of categories with fewer samples, we further analyze the precision and recall of the pseudo-labels for the 10 least-represented categories in the MS-COCO dataset under a 20\% label proportion, as shown in Figure \ref{Fig:least10}. We adopt SST as the main model. Despite the small sample sizes in these categories, the proposed method achieves high precision, with an average of 97.1\%, minimizing the risk of confusion in the model’s training process. Additionally, we obtain an average label recall of 33.2\% for these categories. This is attributed to the low tolerance for false positives in the category thresholds of the proposed method. Moreover, the uncertainty in the distribution of these categories gradually decreases as the model progresses in training, further enhancing both the precision and recall of the pseudo-labels.}

\subsubsection{\textcolor{black}{Analysis of Training Stability}}
\begin{figure}[htbp]
	\centering 
	\includegraphics[width=1.0\linewidth]{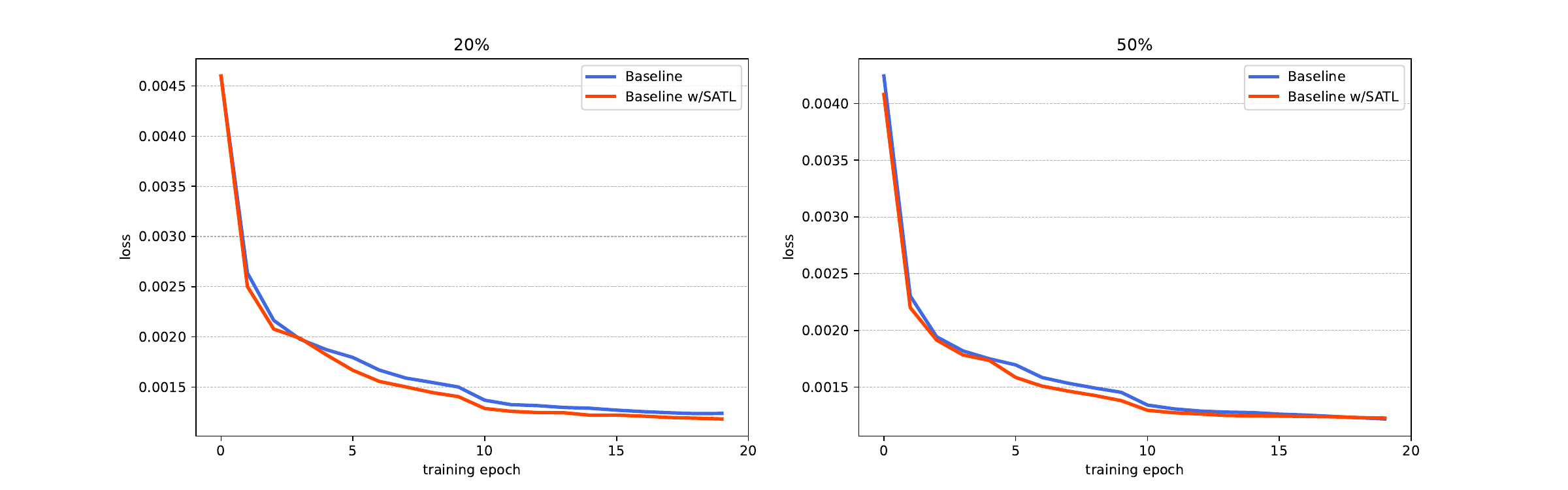}
	\caption{\textcolor{black}{Analysis of the class loss curves on the MS-COCO dataset with 20\% (left) and 50\% (right) known labels. SST is adopted as the baseline model in these experiments.}}
	\label{Fig:lossCurves}
\end{figure}

\textcolor{black}{We further investigate the impact of the proposed SATL module on the training of the main model. To evaluate this, we compare the class loss curves of the model before and after incorporating the SATL module on MS-COCO dataset with 20\% and 50\% known labels, which illustrate the model's training progression, as shown in Figure \ref{Fig:lossCurves}. Despite the increased complexity introduced by the selection of category-specific thresholds, the model maintains a stable training process. Furthermore, by leveraging more high-quality pseudo-labels, the model converges faster in the later stages of training.}

\subsection{\textcolor{black}{Limitations}}
\textcolor{black}{Despite the effectiveness of the proposed SATL method in mining high-precision pseudo-labels, it still exhibits certain limitations, particularly when known and unknown samples are distributed in different domains. In such cases, the prediction distributions for known and unknown samples may become misaligned due to the domain gap, which can be observed in practical applications and leads to improper estimation of category thresholds for the unknown distribution. This motivates us to explore more refined strategies for distribution estimation and threshold selection. Additionally, the proposed method shows relatively lower accuracy for pseudo-labels in certain categories. In future work, we aim to investigate pseudo-label mining and error correction mechanisms that leverage contextual information among categories, providing more accurate training information.}

\section{Conclusion}
In this work, we aim to address the problem of threshold setting in pseudo-labeling methods within the context of MLR-PL task, where only part of labels are known while others are invisible to the model. 
The proposed semantic-aware threshold learning (SATL) framework consists of a semantic-aware threshold estimation module (SATE) and a differential ranking loss function (DRL). Specifically, the SATE assigns class-specific thresholds to different categories based on the output distributions of known samples, considering both the learning status and intrinsic differences between categories. Additionally, the DRL further enhances the discrimination between positive and negative distributions in each category.
These complementary components incorporate to enhance the quantity of recalled labels, and they can be seamlessly incorporated into existing MLR-PL frameworks. To demonstrate the effectiveness of our method, we implement it based on mainstream MLR-PL frameworks. We also conduct comprehensive experiments on various multi-label datasets, including MS-COCO and VG-200, and conduct visualization analysis on our methods for better understanding. These results demonstrate that the proposed SATL method benefits the training process of MLR model through a suitable threshold selection strategy.

\section*{Acknowledgements}
This work was supported in part by National Natural Science Foundation of China (NSFC) under Grant No. 62206060, in Part by Guangdong Basic and Applied Basic Research Foundation (nos. 2025A1515010454, 2023A1515012561).

\bibliographystyle{elsarticle-harv} 
\bibliography{ref}

\begin{thebibliography}{51}
\expandafter\ifx\csname natexlab\endcsname\relax\def\natexlab#1{#1}\fi
\providecommand{\url}[1]{\texttt{#1}}
\providecommand{\href}[2]{#2}
\providecommand{\path}[1]{#1}
\providecommand{\DOIprefix}{doi:}
\providecommand{\ArXivprefix}{arXiv:}
\providecommand{\URLprefix}{URL: }
\providecommand{\Pubmedprefix}{pmid:}
\providecommand{\doi}[1]{\href{http://dx.doi.org/#1}{\path{#1}}}
\providecommand{\Pubmed}[1]{\href{pmid:#1}{\path{#1}}}
\providecommand{\bibinfo}[2]{#2}
\ifx\xfnm\relax \def\xfnm[#1]{\unskip,\space#1}\fi
\bibitem[{Carrillo et~al.(2013)Carrillo, L{\'o}pez and Moreno}]{carrillo2013multi}
\bibinfo{author}{Carrillo, D.}, \bibinfo{author}{L{\'o}pez, V.F.}, \bibinfo{author}{Moreno, M.N.}, \bibinfo{year}{2013}.
\newblock \bibinfo{title}{Multi-label classification for recommender systems}, in: \bibinfo{booktitle}{Trends in Practical Applications of Agents and Multiagent Systems: 11th International Conference on Practical Applications of Agents and Multi-Agent Systems}, \bibinfo{organization}{Springer}. pp. \bibinfo{pages}{181--188}.
\bibitem[{Chen et~al.(2024a)Chen, Lin, Yang, Qing and Lin}]{chen2024learning}
\bibinfo{author}{Chen, T.}, \bibinfo{author}{Lin, J.}, \bibinfo{author}{Yang, Z.}, \bibinfo{author}{Qing, C.}, \bibinfo{author}{Lin, L.}, \bibinfo{year}{2024}a.
\newblock \bibinfo{title}{Learning adaptive spatial coherent correlations for speech-preserving facial expression manipulation}, in: \bibinfo{booktitle}{Proceedings of the IEEE/CVF Conference on Computer Vision and Pattern Recognition}, pp. \bibinfo{pages}{7267--7276}.
\bibitem[{Chen et~al.(2022a)Chen, Lin, Chen, Hui and Wu}]{chen2022knowledge}
\bibinfo{author}{Chen, T.}, \bibinfo{author}{Lin, L.}, \bibinfo{author}{Chen, R.}, \bibinfo{author}{Hui, X.}, \bibinfo{author}{Wu, H.}, \bibinfo{year}{2022}a.
\newblock \bibinfo{title}{Knowledge-guided multi-label few-shot learning for general image recognition}.
\newblock \bibinfo{journal}{IEEE Transactions on Pattern Analysis and Machine Intelligence} \bibinfo{volume}{44}, \bibinfo{pages}{1371--1384}.
\newblock \DOIprefix\doi{10.1109/TPAMI.2020.3025814}.
\bibitem[{Chen et~al.(2024b)Chen, Pu, Liu, Shi, Yang and Lin}]{chen2022heterogeneous}
\bibinfo{author}{Chen, T.}, \bibinfo{author}{Pu, T.}, \bibinfo{author}{Liu, L.}, \bibinfo{author}{Shi, Y.}, \bibinfo{author}{Yang, Z.}, \bibinfo{author}{Lin, L.}, \bibinfo{year}{2024}b.
\newblock \bibinfo{title}{Heterogeneous semantic transfer for multi-label recognition with partial labels}.
\newblock \bibinfo{journal}{International Journal of Computer Vision} , \bibinfo{pages}{6091–6106}.
\bibitem[{Chen et~al.(2022b)Chen, Pu, Wu, Xie and Lin}]{chen2022structured}
\bibinfo{author}{Chen, T.}, \bibinfo{author}{Pu, T.}, \bibinfo{author}{Wu, H.}, \bibinfo{author}{Xie, Y.}, \bibinfo{author}{Lin, L.}, \bibinfo{year}{2022}b.
\newblock \bibinfo{title}{Structured semantic transfer for multi-label recognition with partial labels}, in: \bibinfo{booktitle}{Proceedings of the AAAI conference on artificial intelligence}, pp. \bibinfo{pages}{339--346}.
\bibitem[{Chen et~al.(2021a)Chen, Pu, Wu, Xie, Liu and Lin}]{chen2021cross}
\bibinfo{author}{Chen, T.}, \bibinfo{author}{Pu, T.}, \bibinfo{author}{Wu, H.}, \bibinfo{author}{Xie, Y.}, \bibinfo{author}{Liu, L.}, \bibinfo{author}{Lin, L.}, \bibinfo{year}{2021}a.
\newblock \bibinfo{title}{Cross-domain facial expression recognition: A unified evaluation benchmark and adversarial graph learning}.
\newblock \bibinfo{journal}{IEEE transactions on pattern analysis and machine intelligence} \bibinfo{volume}{44}, \bibinfo{pages}{9887--9903}.
\bibitem[{Chen et~al.(2024c)Chen, Wang, Pu, Qin, Yang, Liu and Lin}]{chen2024dynamic}
\bibinfo{author}{Chen, T.}, \bibinfo{author}{Wang, W.}, \bibinfo{author}{Pu, T.}, \bibinfo{author}{Qin, J.}, \bibinfo{author}{Yang, Z.}, \bibinfo{author}{Liu, J.}, \bibinfo{author}{Lin, L.}, \bibinfo{year}{2024}c.
\newblock \bibinfo{title}{Dynamic correlation learning and regularization for multi-label confidence calibration}.
\newblock \bibinfo{journal}{IEEE Transactions on Image Processing} .
\bibitem[{Chen et~al.(2018)Chen, Wang, Li and Lin}]{chen2018recurrent}
\bibinfo{author}{Chen, T.}, \bibinfo{author}{Wang, Z.}, \bibinfo{author}{Li, G.}, \bibinfo{author}{Lin, L.}, \bibinfo{year}{2018}.
\newblock \bibinfo{title}{Recurrent attentional reinforcement learning for multi-label image recognition}, in: \bibinfo{booktitle}{Proceedings of the AAAI conference on artificial intelligence}.
\bibitem[{Chen et~al.(2019a)Chen, Xu, Hui, Wu and Lin}]{chen2019learning}
\bibinfo{author}{Chen, T.}, \bibinfo{author}{Xu, M.}, \bibinfo{author}{Hui, X.}, \bibinfo{author}{Wu, H.}, \bibinfo{author}{Lin, L.}, \bibinfo{year}{2019}a.
\newblock \bibinfo{title}{Learning semantic-specific graph representation for multi-label image recognition}, in: \bibinfo{booktitle}{Proceedings of the IEEE/CVF international conference on computer vision}, pp. \bibinfo{pages}{522--531}.
\bibitem[{Chen et~al.(2019b)Chen, Wei, Wang and Guo}]{chen2019multi}
\bibinfo{author}{Chen, Z.M.}, \bibinfo{author}{Wei, X.S.}, \bibinfo{author}{Wang, P.}, \bibinfo{author}{Guo, Y.}, \bibinfo{year}{2019}b.
\newblock \bibinfo{title}{Multi-label image recognition with graph convolutional networks}, in: \bibinfo{booktitle}{Proceedings of the IEEE/CVF conference on computer vision and pattern recognition}, pp. \bibinfo{pages}{5177--5186}.
\bibitem[{Chen et~al.(2021b)Chen, Wei, Wang and Guo}]{chen2021learning}
\bibinfo{author}{Chen, Z.M.}, \bibinfo{author}{Wei, X.S.}, \bibinfo{author}{Wang, P.}, \bibinfo{author}{Guo, Y.}, \bibinfo{year}{2021}b.
\newblock \bibinfo{title}{Learning graph convolutional networks for multi-label recognition and applications}.
\newblock \bibinfo{journal}{IEEE Transactions on Pattern Analysis and Machine Intelligence} \bibinfo{volume}{45}, \bibinfo{pages}{6969--6983}.
\bibitem[{Cheng et~al.(2005)Cheng, Chou, Yang and Chang}]{cheng2005semantic}
\bibinfo{author}{Cheng, S.C.}, \bibinfo{author}{Chou, T.C.}, \bibinfo{author}{Yang, C.L.}, \bibinfo{author}{Chang, H.Y.}, \bibinfo{year}{2005}.
\newblock \bibinfo{title}{A semantic learning for content-based image retrieval using analytical hierarchy process}.
\newblock \bibinfo{journal}{Expert Systems with Applications} \bibinfo{volume}{28}, \bibinfo{pages}{495--505}.
\bibitem[{Darban and Valipour(2022)}]{darban2022ghrs}
\bibinfo{author}{Darban, Z.Z.}, \bibinfo{author}{Valipour, M.H.}, \bibinfo{year}{2022}.
\newblock \bibinfo{title}{Ghrs: Graph-based hybrid recommendation system with application to movie recommendation}.
\newblock \bibinfo{journal}{Expert Systems with Applications} \bibinfo{volume}{200}, \bibinfo{pages}{116850}.
\bibitem[{Durand et~al.(2019)Durand, Mehrasa and Mori}]{durand2019learning}
\bibinfo{author}{Durand, T.}, \bibinfo{author}{Mehrasa, N.}, \bibinfo{author}{Mori, G.}, \bibinfo{year}{2019}.
\newblock \bibinfo{title}{Learning a deep convnet for multi-label classification with partial labels}, in: \bibinfo{booktitle}{Proceedings of the IEEE/CVF conference on computer vision and pattern recognition}, pp. \bibinfo{pages}{647--657}.
\bibitem[{Gao and Zhou(2021)}]{gao2021learning}
\bibinfo{author}{Gao, B.B.}, \bibinfo{author}{Zhou, H.Y.}, \bibinfo{year}{2021}.
\newblock \bibinfo{title}{Learning to discover multi-class attentional regions for multi-label image recognition}.
\newblock \bibinfo{journal}{IEEE Transactions on Image Processing} \bibinfo{volume}{30}, \bibinfo{pages}{5920--5932}.
\bibitem[{Huynh and Elhamifar(2020)}]{huynh2020interactive}
\bibinfo{author}{Huynh, D.}, \bibinfo{author}{Elhamifar, E.}, \bibinfo{year}{2020}.
\newblock \bibinfo{title}{Interactive multi-label cnn learning with partial labels}, in: \bibinfo{booktitle}{Proceedings of the IEEE/CVF Conference on Computer Vision and Pattern Recognition}, pp. \bibinfo{pages}{9423--9432}.
\bibitem[{Joulin et~al.(2016)Joulin, Van Der~Maaten, Jabri and Vasilache}]{joulin2016learning}
\bibinfo{author}{Joulin, A.}, \bibinfo{author}{Van Der~Maaten, L.}, \bibinfo{author}{Jabri, A.}, \bibinfo{author}{Vasilache, N.}, \bibinfo{year}{2016}.
\newblock \bibinfo{title}{Learning visual features from large weakly supervised data}, in: \bibinfo{booktitle}{Computer Vision--ECCV 2016: 14th European Conference, Amsterdam, The Netherlands, October 11--14, 2016, Proceedings, Part VII 14}, \bibinfo{organization}{Springer}. pp. \bibinfo{pages}{67--84}.
\bibitem[{Kim et~al.(2022)Kim, Kim, Akata and Lee}]{kim2022large}
\bibinfo{author}{Kim, Y.}, \bibinfo{author}{Kim, J.M.}, \bibinfo{author}{Akata, Z.}, \bibinfo{author}{Lee, J.}, \bibinfo{year}{2022}.
\newblock \bibinfo{title}{Large loss matters in weakly supervised multi-label classification}, in: \bibinfo{booktitle}{Proceedings of the IEEE/CVF Conference on Computer Vision and Pattern Recognition}, pp. \bibinfo{pages}{14156--14165}.
\bibitem[{Krishna et~al.(2017)Krishna, Zhu, Groth, Johnson, Hata, Kravitz, Chen, Kalantidis, Li, Shamma et~al.}]{krishna2017visual}
\bibinfo{author}{Krishna, R.}, \bibinfo{author}{Zhu, Y.}, \bibinfo{author}{Groth, O.}, \bibinfo{author}{Johnson, J.}, \bibinfo{author}{Hata, K.}, \bibinfo{author}{Kravitz, J.}, \bibinfo{author}{Chen, S.}, \bibinfo{author}{Kalantidis, Y.}, \bibinfo{author}{Li, L.J.}, \bibinfo{author}{Shamma, D.A.}, et~al., \bibinfo{year}{2017}.
\newblock \bibinfo{title}{Visual genome: Connecting language and vision using crowdsourced dense image annotations}.
\newblock \bibinfo{journal}{International journal of computer vision} \bibinfo{volume}{123}, \bibinfo{pages}{32--73}.
\bibitem[{Kundu and Tighe(2020)}]{kundu2020exploiting}
\bibinfo{author}{Kundu, K.}, \bibinfo{author}{Tighe, J.}, \bibinfo{year}{2020}.
\newblock \bibinfo{title}{Exploiting weakly supervised visual patterns to learn from partial annotations}.
\newblock \bibinfo{journal}{Advances in Neural Information Processing Systems} \bibinfo{volume}{33}, \bibinfo{pages}{561--572}.
\bibitem[{Lai et~al.(2016)Lai, Yan, Shu, Wei and Yan}]{lai2016instance}
\bibinfo{author}{Lai, H.}, \bibinfo{author}{Yan, P.}, \bibinfo{author}{Shu, X.}, \bibinfo{author}{Wei, Y.}, \bibinfo{author}{Yan, S.}, \bibinfo{year}{2016}.
\newblock \bibinfo{title}{Instance-aware hashing for multi-label image retrieval}.
\newblock \bibinfo{journal}{IEEE Transactions on Image Processing} \bibinfo{volume}{25}, \bibinfo{pages}{2469--2479}.
\bibitem[{Li et~al.(2022)Li, Wu, Shrivastava and Davis}]{li2022rethinking}
\bibinfo{author}{Li, H.}, \bibinfo{author}{Wu, Z.}, \bibinfo{author}{Shrivastava, A.}, \bibinfo{author}{Davis, L.S.}, \bibinfo{year}{2022}.
\newblock \bibinfo{title}{Rethinking pseudo labels for semi-supervised object detection}, in: \bibinfo{booktitle}{Proceedings of the AAAI Conference on Artificial Intelligence}, pp. \bibinfo{pages}{1314--1322}.
\bibitem[{Li et~al.(2010)Li, Zhang, Lu, Lu and Tian}]{li2010technique}
\bibinfo{author}{Li, R.}, \bibinfo{author}{Zhang, Y.}, \bibinfo{author}{Lu, Z.}, \bibinfo{author}{Lu, J.}, \bibinfo{author}{Tian, Y.}, \bibinfo{year}{2010}.
\newblock \bibinfo{title}{Technique of image retrieval based on multi-label image annotation}, in: \bibinfo{booktitle}{2010 Second International Conference on Multimedia and Information Technology}, \bibinfo{organization}{IEEE}. pp. \bibinfo{pages}{10--13}.
\bibitem[{Lima and De~Castro(2014)}]{lima2014multi}
\bibinfo{author}{Lima, A.C.E.}, \bibinfo{author}{De~Castro, L.N.}, \bibinfo{year}{2014}.
\newblock \bibinfo{title}{A multi-label, semi-supervised classification approach applied to personality prediction in social media}.
\newblock \bibinfo{journal}{Neural Networks} \bibinfo{volume}{58}, \bibinfo{pages}{122--130}.
\bibitem[{Lin et~al.(2014)Lin, Maire, Belongie, Hays, Perona, Ramanan, Doll{\'a}r and Zitnick}]{lin2014microsoft}
\bibinfo{author}{Lin, T.Y.}, \bibinfo{author}{Maire, M.}, \bibinfo{author}{Belongie, S.}, \bibinfo{author}{Hays, J.}, \bibinfo{author}{Perona, P.}, \bibinfo{author}{Ramanan, D.}, \bibinfo{author}{Doll{\'a}r, P.}, \bibinfo{author}{Zitnick, C.L.}, \bibinfo{year}{2014}.
\newblock \bibinfo{title}{Microsoft coco: Common objects in context}, in: \bibinfo{booktitle}{Computer Vision--ECCV 2014: 13th European Conference, Zurich, Switzerland, September 6-12, 2014, Proceedings, Part V 13}, \bibinfo{organization}{Springer}. pp. \bibinfo{pages}{740--755}.
\bibitem[{Ma and Chow(2019)}]{ma2019label}
\bibinfo{author}{Ma, J.}, \bibinfo{author}{Chow, T.W.}, \bibinfo{year}{2019}.
\newblock \bibinfo{title}{Label-specific feature selection and two-level label recovery for multi-label classification with missing labels}.
\newblock \bibinfo{journal}{Neural Networks} \bibinfo{volume}{118}, \bibinfo{pages}{110--126}.
\bibitem[{Ou et~al.(2020)Ou, Yu, Domeniconi, Lu and Zhang}]{ou2020multi}
\bibinfo{author}{Ou, G.}, \bibinfo{author}{Yu, G.}, \bibinfo{author}{Domeniconi, C.}, \bibinfo{author}{Lu, X.}, \bibinfo{author}{Zhang, X.}, \bibinfo{year}{2020}.
\newblock \bibinfo{title}{Multi-label zero-shot learning with graph convolutional networks}.
\newblock \bibinfo{journal}{Neural Networks} \bibinfo{volume}{132}, \bibinfo{pages}{333--341}.
\bibitem[{Pu et~al.(2024)Pu, Chen, Wu, Shi, Yang and Lin}]{pu2024dual}
\bibinfo{author}{Pu, T.}, \bibinfo{author}{Chen, T.}, \bibinfo{author}{Wu, H.}, \bibinfo{author}{Shi, Y.}, \bibinfo{author}{Yang, Z.}, \bibinfo{author}{Lin, L.}, \bibinfo{year}{2024}.
\newblock \bibinfo{title}{Dual-perspective semantic-aware representation blending for multi-label image recognition with partial labels}.
\newblock \bibinfo{journal}{Expert Systems with Applications} , \bibinfo{pages}{123526}.
\bibitem[{Pu et~al.(2021)Pu, Chen, Xie, Wu and Lin}]{pu2021expression}
\bibinfo{author}{Pu, T.}, \bibinfo{author}{Chen, T.}, \bibinfo{author}{Xie, Y.}, \bibinfo{author}{Wu, H.}, \bibinfo{author}{Lin, L.}, \bibinfo{year}{2021}.
\newblock \bibinfo{title}{Au-expression knowledge constrained representation learning for facial expression recognition}, in: \bibinfo{booktitle}{2021 IEEE international conference on robotics and automation (ICRA)}, \bibinfo{organization}{IEEE}. pp. \bibinfo{pages}{11154--11161}.
\bibitem[{Ridnik et~al.(2021)Ridnik, Ben-Baruch, Zamir, Noy, Friedman, Protter and Zelnik-Manor}]{ridnik2021asymmetric}
\bibinfo{author}{Ridnik, T.}, \bibinfo{author}{Ben-Baruch, E.}, \bibinfo{author}{Zamir, N.}, \bibinfo{author}{Noy, A.}, \bibinfo{author}{Friedman, I.}, \bibinfo{author}{Protter, M.}, \bibinfo{author}{Zelnik-Manor, L.}, \bibinfo{year}{2021}.
\newblock \bibinfo{title}{Asymmetric loss for multi-label classification}, in: \bibinfo{booktitle}{Proceedings of the IEEE/CVF international conference on computer vision}, pp. \bibinfo{pages}{82--91}.
\bibitem[{Rizve et~al.(2021)Rizve, Duarte, Rawat and Shah}]{rizve2021defense}
\bibinfo{author}{Rizve, M.N.}, \bibinfo{author}{Duarte, K.}, \bibinfo{author}{Rawat, Y.S.}, \bibinfo{author}{Shah, M.}, \bibinfo{year}{2021}.
\newblock \bibinfo{title}{In defense of pseudo-labeling: An uncertainty-aware pseudo-label selection framework for semi-supervised learning}.
\newblock \bibinfo{journal}{arXiv preprint arXiv:2101.06329} .
\bibitem[{Ruan et~al.(2024)Ruan, Xu, Yang, Lu, Qin and Chen}]{ruan2024learning}
\bibinfo{author}{Ruan, H.}, \bibinfo{author}{Xu, Z.}, \bibinfo{author}{Yang, Z.}, \bibinfo{author}{Lu, Y.}, \bibinfo{author}{Qin, J.}, \bibinfo{author}{Chen, T.}, \bibinfo{year}{2024}.
\newblock \bibinfo{title}{Learning semantic-aware representation in visual-language models for multi-label recognition with partial labels}.
\newblock \bibinfo{journal}{ACM Trans. Multimedia Comput. Commun. Appl.} .
\bibitem[{Sohn et~al.(2020)Sohn, Berthelot, Carlini, Zhang, Zhang, Raffel, Cubuk, Kurakin and Li}]{sohn2020fixmatch}
\bibinfo{author}{Sohn, K.}, \bibinfo{author}{Berthelot, D.}, \bibinfo{author}{Carlini, N.}, \bibinfo{author}{Zhang, Z.}, \bibinfo{author}{Zhang, H.}, \bibinfo{author}{Raffel, C.A.}, \bibinfo{author}{Cubuk, E.D.}, \bibinfo{author}{Kurakin, A.}, \bibinfo{author}{Li, C.L.}, \bibinfo{year}{2020}.
\newblock \bibinfo{title}{Fixmatch: Simplifying semi-supervised learning with consistency and confidence}.
\newblock \bibinfo{journal}{Advances in neural information processing systems} \bibinfo{volume}{33}, \bibinfo{pages}{596--608}.
\bibitem[{Sun et~al.(2017)Sun, Shrivastava, Singh and Gupta}]{sun2017revisiting}
\bibinfo{author}{Sun, C.}, \bibinfo{author}{Shrivastava, A.}, \bibinfo{author}{Singh, S.}, \bibinfo{author}{Gupta, A.}, \bibinfo{year}{2017}.
\newblock \bibinfo{title}{Revisiting unreasonable effectiveness of data in deep learning era}, in: \bibinfo{booktitle}{Proceedings of the IEEE international conference on computer vision}, pp. \bibinfo{pages}{843--852}.
\bibitem[{Sun et~al.(2014)Sun, Tang, Li, Qi and Huang}]{sun2014multi}
\bibinfo{author}{Sun, F.}, \bibinfo{author}{Tang, J.}, \bibinfo{author}{Li, H.}, \bibinfo{author}{Qi, G.J.}, \bibinfo{author}{Huang, T.S.}, \bibinfo{year}{2014}.
\newblock \bibinfo{title}{Multi-label image categorization with sparse factor representation}.
\newblock \bibinfo{journal}{IEEE Transactions on Image Processing} \bibinfo{volume}{23}, \bibinfo{pages}{1028--1037}.
\bibitem[{Tang et~al.(2022)Tang, Li and Tian}]{tang2022image}
\bibinfo{author}{Tang, J.}, \bibinfo{author}{Li, D.}, \bibinfo{author}{Tian, Y.}, \bibinfo{year}{2022}.
\newblock \bibinfo{title}{Image classification with multi-view multi-instance metric learning}.
\newblock \bibinfo{journal}{Expert Systems with Applications} \bibinfo{volume}{189}, \bibinfo{pages}{116117}.
\bibitem[{Tian et~al.(2023)Tian, Bai, Yu and Zhu}]{tian2023causal}
\bibinfo{author}{Tian, Y.}, \bibinfo{author}{Bai, K.}, \bibinfo{author}{Yu, X.}, \bibinfo{author}{Zhu, S.}, \bibinfo{year}{2023}.
\newblock \bibinfo{title}{Causal multi-label learning for image classification}.
\newblock \bibinfo{journal}{Neural Networks} \bibinfo{volume}{167}, \bibinfo{pages}{626--637}.
\bibitem[{Tsoumakas and Katakis(2007)}]{tsoumakas2007multi}
\bibinfo{author}{Tsoumakas, G.}, \bibinfo{author}{Katakis, I.}, \bibinfo{year}{2007}.
\newblock \bibinfo{title}{Multi-label classification: An overview}.
\newblock \bibinfo{journal}{International Journal of Data Warehousing and Mining (IJDWM)} \bibinfo{volume}{3}, \bibinfo{pages}{1--13}.
\bibitem[{Wang et~al.(2021)Wang, Xiao, Li, Feng, Niu, Chen and Zhao}]{wang2021pico}
\bibinfo{author}{Wang, H.}, \bibinfo{author}{Xiao, R.}, \bibinfo{author}{Li, Y.}, \bibinfo{author}{Feng, L.}, \bibinfo{author}{Niu, G.}, \bibinfo{author}{Chen, G.}, \bibinfo{author}{Zhao, J.}, \bibinfo{year}{2021}.
\newblock \bibinfo{title}{Pico: Contrastive label disambiguation for partial label learning}, in: \bibinfo{booktitle}{International Conference on Learning Representations}.
\bibitem[{Wang et~al.(2016)Wang, Yang, Mao, Huang, Huang and Xu}]{wang2016cnn}
\bibinfo{author}{Wang, J.}, \bibinfo{author}{Yang, Y.}, \bibinfo{author}{Mao, J.}, \bibinfo{author}{Huang, Z.}, \bibinfo{author}{Huang, C.}, \bibinfo{author}{Xu, W.}, \bibinfo{year}{2016}.
\newblock \bibinfo{title}{Cnn-rnn: A unified framework for multi-label image classification}, in: \bibinfo{booktitle}{Proceedings of the IEEE conference on computer vision and pattern recognition}, pp. \bibinfo{pages}{2285--2294}.
\bibitem[{Wang et~al.(2023)Wang, Zhao, Wang, Xu and Sun}]{wang2023image}
\bibinfo{author}{Wang, M.}, \bibinfo{author}{Zhao, Y.}, \bibinfo{author}{Wang, Y.}, \bibinfo{author}{Xu, T.}, \bibinfo{author}{Sun, Y.}, \bibinfo{year}{2023}.
\newblock \bibinfo{title}{Image emotion multi-label classification based on multi-graph learning}.
\newblock \bibinfo{journal}{Expert Systems with Applications} \bibinfo{volume}{231}, \bibinfo{pages}{120641}.
\bibitem[{Wang et~al.(2017)Wang, Chen, Li, Xu and Lin}]{wang2017multi}
\bibinfo{author}{Wang, Z.}, \bibinfo{author}{Chen, T.}, \bibinfo{author}{Li, G.}, \bibinfo{author}{Xu, R.}, \bibinfo{author}{Lin, L.}, \bibinfo{year}{2017}.
\newblock \bibinfo{title}{Multi-label image recognition by recurrently discovering attentional regions}, in: \bibinfo{booktitle}{Proceedings of the IEEE international conference on computer vision}, pp. \bibinfo{pages}{464--472}.
\bibitem[{Wei et~al.(2015)Wei, Xia, Lin, Huang, Ni, Dong, Zhao and Yan}]{wei2015hcp}
\bibinfo{author}{Wei, Y.}, \bibinfo{author}{Xia, W.}, \bibinfo{author}{Lin, M.}, \bibinfo{author}{Huang, J.}, \bibinfo{author}{Ni, B.}, \bibinfo{author}{Dong, J.}, \bibinfo{author}{Zhao, Y.}, \bibinfo{author}{Yan, S.}, \bibinfo{year}{2015}.
\newblock \bibinfo{title}{Hcp: A flexible cnn framework for multi-label image classification}.
\newblock \bibinfo{journal}{IEEE transactions on pattern analysis and machine intelligence} \bibinfo{volume}{38}, \bibinfo{pages}{1901--1907}.
\bibitem[{Wu et~al.(2019)Wu, Hu, Wang, Li, Nie and Cheng}]{wu2019instance}
\bibinfo{author}{Wu, H.}, \bibinfo{author}{Hu, Y.}, \bibinfo{author}{Wang, K.}, \bibinfo{author}{Li, H.}, \bibinfo{author}{Nie, L.}, \bibinfo{author}{Cheng, H.}, \bibinfo{year}{2019}.
\newblock \bibinfo{title}{Instance-aware representation learning and association for online multi-person tracking}.
\newblock \bibinfo{journal}{Pattern Recognition} \bibinfo{volume}{94}, \bibinfo{pages}{25--34}.
\bibitem[{Wu et~al.(2020)Wu, Chen, Li, Xiao and Hu}]{wu2020adahgnn}
\bibinfo{author}{Wu, X.}, \bibinfo{author}{Chen, Q.}, \bibinfo{author}{Li, W.}, \bibinfo{author}{Xiao, Y.}, \bibinfo{author}{Hu, B.}, \bibinfo{year}{2020}.
\newblock \bibinfo{title}{Adahgnn: Adaptive hypergraph neural networks for multi-label image classification}, in: \bibinfo{booktitle}{Proceedings of the 28th ACM International Conference on Multimedia}, pp. \bibinfo{pages}{284--293}.
\bibitem[{Yang et~al.(2016)Yang, Tianyi~Zhou, Zhang, Gao, Wu and Cai}]{yang2016exploit}
\bibinfo{author}{Yang, H.}, \bibinfo{author}{Tianyi~Zhou, J.}, \bibinfo{author}{Zhang, Y.}, \bibinfo{author}{Gao, B.B.}, \bibinfo{author}{Wu, J.}, \bibinfo{author}{Cai, J.}, \bibinfo{year}{2016}.
\newblock \bibinfo{title}{Exploit bounding box annotations for multi-label object recognition}, in: \bibinfo{booktitle}{Proceedings of the IEEE conference on computer vision and pattern recognition}, pp. \bibinfo{pages}{280--288}.
\bibitem[{Zhang et~al.(2021)Zhang, Wang, Hou, Wu, Wang, Okumura and Shinozaki}]{zhang2021flexmatch}
\bibinfo{author}{Zhang, B.}, \bibinfo{author}{Wang, Y.}, \bibinfo{author}{Hou, W.}, \bibinfo{author}{Wu, H.}, \bibinfo{author}{Wang, J.}, \bibinfo{author}{Okumura, M.}, \bibinfo{author}{Shinozaki, T.}, \bibinfo{year}{2021}.
\newblock \bibinfo{title}{Flexmatch: Boosting semi-supervised learning with curriculum pseudo labeling}.
\newblock \bibinfo{journal}{Advances in Neural Information Processing Systems} \bibinfo{volume}{34}, \bibinfo{pages}{18408--18419}.
\bibitem[{Zhang et~al.(2023a)Zhang, Meng, Cao, Liu, Ming and Yang}]{zhang2023graph}
\bibinfo{author}{Zhang, H.}, \bibinfo{author}{Meng, X.}, \bibinfo{author}{Cao, W.}, \bibinfo{author}{Liu, Y.}, \bibinfo{author}{Ming, Z.}, \bibinfo{author}{Yang, J.}, \bibinfo{year}{2023}a.
\newblock \bibinfo{title}{Graph embedding based multi-label zero-shot learning}.
\newblock \bibinfo{journal}{Neural Networks} \bibinfo{volume}{167}, \bibinfo{pages}{129--140}.
\bibitem[{Zhang et~al.(2023b)Zhang, Zhang and Liu}]{zhang2023refined}
\bibinfo{author}{Zhang, S.}, \bibinfo{author}{Zhang, L.}, \bibinfo{author}{Liu, Z.}, \bibinfo{year}{2023}b.
\newblock \bibinfo{title}{Refined pseudo labeling for source-free domain adaptive object detection}, in: \bibinfo{booktitle}{ICASSP 2023-2023 IEEE International Conference on Acoustics, Speech and Signal Processing (ICASSP)}, \bibinfo{organization}{IEEE}. pp. \bibinfo{pages}{1--5}.
\bibitem[{Zhang and Peng(2021)}]{zhang2021instance}
\bibinfo{author}{Zhang, Z.}, \bibinfo{author}{Peng, H.}, \bibinfo{year}{2021}.
\newblock \bibinfo{title}{Instance-weighted central similarity for multi-label image retrieval}.
\newblock \bibinfo{journal}{arXiv preprint arXiv:2108.05274} .
\bibitem[{Zheng et~al.(2014)Zheng, Mobasher and Burke}]{zheng2014context}
\bibinfo{author}{Zheng, Y.}, \bibinfo{author}{Mobasher, B.}, \bibinfo{author}{Burke, R.}, \bibinfo{year}{2014}.
\newblock \bibinfo{title}{Context recommendation using multi-label classification}, in: \bibinfo{booktitle}{2014 IEEE/WIC/ACM International Joint Conferences on Web Intelligence (WI) and Intelligent Agent Technologies (IAT)}, \bibinfo{organization}{IEEE}. pp. \bibinfo{pages}{288--295}.

\end{thebibliography}






\end{document}